\documentclass[11pt,a4paper]{article}

\usepackage[T1]{fontenc}
\usepackage[utf8]{inputenc}
\usepackage{lmodern}
\usepackage[margin=1in]{geometry}
\usepackage{graphicx}
\usepackage{amsmath,amssymb}
\usepackage{booktabs}
\usepackage{tabularx}
\usepackage{caption}
\usepackage{authblk}
\usepackage[numbers,sort&compress]{natbib}
\usepackage{microtype}
\usepackage{float}
\usepackage{placeins}
\usepackage{enumerate}
\usepackage{xcolor}
\usepackage{multirow}
\usepackage{chngcntr}
\usepackage[colorlinks=true,allcolors=blue!55!black]{hyperref}

\newcommand{\lam}{\lambda}
\newcommand{\ci}[2]{(95\% CI #1, #2)}

\newif\ifhighlightrevisions
\highlightrevisionsfalse
\ifhighlightrevisions
  \newcommand{\rev}[1]{{\color{red}#1}}
\else
  \newcommand{\rev}[1]{#1}
\fi

\usepackage{authblk}
\usepackage{hyperref}

\title{Payoff scaling shapes cooperation in LLM agents across languages}

\author[]{%
The Anh Han\textsuperscript{1\textdagger*},
Trung-Kiet Huynh\textsuperscript{2,4\textdagger},
Dao-Sy Duy-Minh\textsuperscript{2,4\textdagger},
Thanh-Bang Cao\textsuperscript{3,4},
Phong-Hao Le\textsuperscript{3,4},
Hong-Dan Nguyen\textsuperscript{3,4},
Phu-Quy Nguyen-Lam\textsuperscript{2,4},
Minh-Luan Nguyen-Vo\textsuperscript{3,4},
Hong-Phat Pham\textsuperscript{3,4},
Phu-Hoa Pham\textsuperscript{2,4},
Thien-Kim Than\textsuperscript{3,4},
Chi-Nguyen Tran\textsuperscript{3,4},
Huy Tran\textsuperscript{3,4},
Gia-Thoai Tran-Le\textsuperscript{3,4},
Alessio Buscemi\textsuperscript{5},
and Le Hong Trang\textsuperscript{3,4*}
}

\affil[]{%
\textsuperscript{1}School of Computing, Engineering and Digital Technologies,
Teesside University, Middlesbrough, United Kingdom\\
\textsuperscript{2}Faculty of Information Technology,
University of Science (HCMUS), Ho Chi Minh City, Vietnam\\
\textsuperscript{3}Faculty of Computer Science and Engineering,
Ho Chi Minh City University of Technology (HCMUT),
Ho Chi Minh City, Vietnam\\
\textsuperscript{4}Vietnam National University,
Ho Chi Minh City (VNU-HCM), Ho Chi Minh City, Vietnam\\
\textsuperscript{5}Luxembourg Institute of Science and Technology (LIST),
Luxembourg
}

\date{%
\vspace{-1.0em}
\footnotesize
Author accepted manuscript. Accepted for publication in \textit{Interface Focus} (Royal Society).\\[2pt]
\textsuperscript{\textdagger}\,These authors contributed equally to this work.\\
\textsuperscript{*}\,Corresponding authors:
\href{mailto:t.han@tees.ac.uk}{t.han@tees.ac.uk}
and
\href{mailto:lhtrang@hcmut.edu.vn}{lhtrang@hcmut.edu.vn}.
}

\begin{document}

\maketitle

\begin{abstract}
Large language models (LLMs) are increasingly deployed as autonomous agents that negotiate, coordinate, and make decisions on behalf of human users. Understanding how they cooperate is therefore important for AI governance and machine behaviour research. We examine two factors: payoff magnitude and interaction language. Classical game theory predicts that multiplying all payoffs by a positive constant should not change behaviour because the underlying strategic incentives remain unchanged. We test this prediction using five widely adopted commercial LLMs playing an iterated prisoner’s dilemma across ten payoff scales and five languages \rev{(English, French, Vietnamese, Chinese and Arabic)}. Contrary to theoretical expectations, changing only the numerical scale of payoffs substantially alters cooperation. Models respond differently, often in opposite directions, meaning that averages across models can conceal important behavioural heterogeneity. Sensitivity to payoff scale also depends strongly on prompt language. Moreover, payoff scale modifies the effects of persona instructions: at sub-unit stakes\rev{, where no payoff shown exceeds 1,} all five models cooperate more under a selfish than a cooperative persona, whereas Claude and Gemini reverse this pattern at larger scales. These findings show that cooperation measures depend critically on payoff scale and language and should not be treated as fixed model traits.
\end{abstract}


\section{Introduction}\label{sec:intro}

Large language models \rev{(LLMs)} are increasingly deployed as agents that negotiate, coordinate and decide on behalf of their users, in recommendation systems, negotiation tools and multi-agent assistants~\cite{park2023generative,tessler2024ai,hammond2025multi,cook2024llm}. In these settings, behaviour emerges from strategic interaction rather than from isolated model outputs, and the agents face cooperation dilemmas with solutions  directly relevant to  safety, coordination and AI governance~\cite{dafoe2021cooperative,lu2024llms,han2026social,idowu2026mapping}. \rev{More broadly, LLM-driven generative agents have been proposed as a new route for studying complex systems and evolutionary games~\cite{lu2024llms,han2025llms,chen2025harnessing}.} A fast-growing literature measures how they behave when they do, using methods from behavioural economics and cognitive psychology~\cite{binz2023using,horton2023large,aher2023using,rilla2026recognising} and  the classical games~\cite{mei2024turing}. Equally consequential is how far that behaviour adapts to the costs and benefits at issue, since a deployed agent meets the same dilemma at many different magnitudes~\cite{hammond2025multi}.

A growing body of work evaluates language models through game-theoretic lenses, using evolutionary and repeated games, reporting systematic departures from equilibrium outcomes, persistent cooperative biases and sensitivity to contextual factors such as language and incentives~\cite{buscemi2025fairgame,sun2025gtllm_survey,mao2025alympics,pal2026strategiescooperationdefectionlarge,willis2025will,fan2024can,pires2025large,fu2026optimal,xie2026evolutionary}. The prisoner's dilemma has been a well-established framework ~\cite{axelrod1981evolution,axelrod1984evolution,nowak1992tit,nowak1993strategy,nowak2006five,pressDyson2012}. Our study is closest to FAIRGAME, which introduced a game-theoretic evaluation framework and reported payoff and language sensitivity in repeated games~\cite{buscemi2025fairgame}. We retain its no-communication prisoner's dilemma protocol so that the comparison is anchored to the same task, while adding a positive-rescaling null spanning ten scales, five languages and five model endpoints. This makes our contribution a test of invariance and heterogeneity, rather than an evaluation of LLM behaviours in a new strategic interaction.

Prior work showed that the LLM models differ in how readily they retaliate~\cite{akata2025} or forgive~\cite{fontana2025nicer}, how much weight they give the game rather than its framing  ~\cite{loreheydari2024strategic} and how far an instructed disposition carries into conditioned reciprocity~\cite{phelps2023investigating}. These studies, however, share two largely unremarked reporting conventions. First, they characterise a model using a cooperation rate, measured under a single payoff matrix. Second, they report what agents did, rather than the decision rule that generated the observed behaviour. This makes it difficult to tell whether a reported cooperation rate is a stable property of the model, an artefact of the particular payoff scale and framing, or the aggregate trace of a different strategic rule.

Evaluating strategic behaviour at the level required for alignment and governance therefore requires more than aggregate outcomes. Drawing on behavioural and evolutionary game theory~\cite{sigmund:2010bo}, a behavioural intention can be conceptualised as an agent's \emph{strategy}: a decision rule mapping an interaction history to the next action~\cite{han2013intention,diStefanoIntention2023,Han2012ALIFEjournal,fujimoto2019functional}. The canonical strategies of repeated games, e.g. Always Cooperate, Always Defect, Tit-for-Tat and Win-Stay-Lose-Shift~\cite{Axelrod1980Effective,sigmund:2010bo}, provide an interpretable vocabulary for such intentions, and prior work has shown that they can be inferred from noisy behavioural trajectories by supervised learning~\cite{han2013intention,diStefanoIntention2023} and extended by probabilistic models to mixed or stochastic play~\cite{inferenceStrategies}. The inference is not free. Separating the memory-one rules from observed play has been shown to need a long run of observations even where the resident rule is fixed by construction~\cite{kim2021win,hilbe2013memory}, which is why the read-out used here is reported as a resemblance, with its provenance beside it (Section \ref{sec:readout}).

In parallel, a language model's behaviour varies with the language in which the prompt is written ~\cite{buscemi2025fairgame}. Strategic reasoning, risk sensitivity and cooperation patterns depend on linguistic framing, sometimes by margins comparable to the differences between models~\cite{loreheydari2024strategic,buscemi2025llms,buscemi2025strategic,powell2026lost}. Yet it remains unclear whether the strategies that agents appear to play are themselves language-dependent, and whether such dependence introduces systematic bias into multi-agent settings. \rev{Language may also change the effect of the stakes, not only the level of cooperation: the effect of stake size on human play has been measured in different societies, with mixed results~\cite{bosman2017emotion,cameron1999raising}, and the cultural values that language models express shift with the prompt~\cite{tao2024cultural}, so a change of scale may be read differently in each language.}

This dependence is an instance of a more general problem: measurements of language-model behaviour are often sensitive to aspects of the elicitation procedure that are not themselves the intended object of measurement. Accuracy can vary with the ordering of in-context examples~\cite{lu2022fantastically}, and with formatting choices that preserve the semantic content of the prompt, by margins large enough to obscure the differences between the models under comparison~\cite{sclar2024quantifying}. Persona instructions are similarly consequential: they alter model outputs~\cite{deshpande2023toxicity}, affect measured capability~\cite{salewski2023context}, and, when specified through a demographic backstory, can induce response distributions resembling those of the corresponding human subgroup~\cite{argyle2023out}. They therefore do not establish a well-defined null condition against which the effect of prompting can be assessed.

The choice of units (of measurement)  is rarely presented as a design decision, because in the standard theory it is not one. \rev{When the entries of a payoff matrix are read as von Neumann-Morgenstern utilities, the preferences they represent are unchanged by any positive affine transformation}~\cite{von2007theory,luce1957games}. Multiplying every entry in the payoff matrix by ten changes no preference ordering, no best response, no equilibrium, and no \rev{orbit} of the replicator dynamics\rev{; it rescales only time, so convergence speeds up or slows down along the same path}. The result is not a new game with larger payoffs, but the same game written in different units.  The theory of the iterated dilemma inherits this invariance: its predictions depend on the ordering of the  payoff-matrix entries  and on ratios among them, rather than on their absolute magnitudes~\cite{axelrod1981evolution,nowak1992tit,nowak1993strategy,pressDyson2012}. For this reason, the modelling literature often normalises the payoff scale away~\cite{szabo2007evolutionary,perc2010coevolutionary,wang2015universal}. \rev{Recent} theory strengthens this point: in models in which social learners update on the payoff they most recently received, the condition for conditional cooperation to persist under strong selection contains neither the benefit nor the cost of cooperation, and depends only on the ordinal structure of the payoffs~\cite{glynatsi2024evolution}.

The invariance is a property of the theory, not of human decision-making. Human valuation is neither linear in magnitude nor anchored at zero~\cite{kahneman1979prospect}, human choice reverses between logically equivalent descriptions of one problem~\cite{tversky1981framing}, and human cooperation in the dilemma itself responds to the size of the stakes~\cite{krockow2016cooperation,list2006friend}. \rev{These effects are usually modest and depend on the task and on stakes that are large relative to income~\cite{camerer1999effects,kocher2008does,engel2011dictator,larney2019stake,cameron1999raising,andersen2011stakes,holt2002risk}.} That gap has been argued to be structural rather than incidental, with the linguistic description of an action entering the utility function directly and a shift proposed from outcome-based to language-based preferences~\cite{capraro2024language}.

Two things are established, and they have not been put together. The first is that what a language model does depends on features of the prompt that carry no task content: the order of the examples, the formatting, the persona, the language. The second is that multiplying every payoff-matrix entry by a positive constant carries no task content in the strictest sense available, because the matrix it produces is  the same game. What is not known is whether the strategic behaviour of a language model respects that invariance. To our knowledge, no prior study has formulated positive payoff rescaling as a theorem-backed zero-effect invariance test across a broad multi-model, multilingual grid. Every manipulation in the sensitivity literature could, in principle, induce legitimate behavioural changes. Consequently, observed changes under these manipulations are difficult to interpret, since the expected direction or magnitude of the effect is typically unspecified. Such changes may therefore be treated either as methodological noise or as empirical findings, but only relative to an implicit and under-theorised baseline. Payoff rescaling is distinctive because its theoretical prediction is explicit: behaviour should remain unchanged. A measured departure from this prediction therefore has a clearly defined null reference, rather than only a contrast with another experimental condition.

Taken together, these considerations motivate three questions about the effects of payoff rescaling:
\begin{enumerate}[(1)]
\item Is strategic behaviour invariant to   rescaling of payoffs, and how large are any deviations from invariance?
\item Does the result depend on the language of the prompt and on the persona assigned to the model?

\item Does payoff rescaling affect only how often the model  cooperates, or does it also alter the underlying strategy it adopts?

\end{enumerate}
We systematically address these questions on a complete and exactly balanced grid of 10{,}000 configurations (dyads) (Section \ref{sec:methodology}), and compare the responses against an evolutionary game baseline computed on the same grid.

Our main contributions  are: \textbf{(i)} A payoff-rescaling protocol whose predicted behavioural effect is exactly zero, demonstrating that five widely adopted commercial language models systematically violate scale invariance (Section \ref{sec:payoff_magnitude}), with cooperation rates shifting substantially across payoff scales. \textbf{(ii)} A per-model analysis showing that scale departures diverge sharply across models to the point of anti-correlation. Consequently, aggregate analyses produce a synthetic average that is representative of no individual model. Variance is dominated by model identity and model-by-scale interactions; while the common scale effect in the five-model decomposition is largely driven by Grok, the aggregate dyad-level association persists in leave-one-model-out analyses. \textbf{(iii)} Cross-lingual and structural evidence demonstrating that payoff-scale sensitivity is strongly language-dependent, modulates persona-induced behaviour at sub-unit stakes, and alters the strategic profiles exhibited by the models. \textbf{(iv)} An analytical evolutionary game baseline where evolutionary outcomes diverge from the aggregate empirical patterns, and a transition analysis demonstrating  that own-action  conditioning generally outweighs partner-action conditioning (Section \ref{sec:conditioning}, appendix~D.7).

\section{Material and methods}\label{sec:methodology}

\subsection{Framework}\label{sec:background}

FAIRGAME, the Framework for AI Agents Bias Recognition using Game Theory~\cite{buscemi2025fairgame}, supplies the computational infrastructure for this study. It supports systematic and reproducible evaluation of language models through controlled game-theoretic experiments: an experimental condition is declared in a configuration file that fixes the payoff structure, the horizon, the model backend and the prompt language; at run time the framework combines that declaration with language-specific prompt templates, simulates the repeated normal-form game, and logs the round-by-round trajectory. We extend it in three ways. First, with a payoff-scaling factor applied to the prisoner's dilemma, which is the manipulation this paper examines (Section \ref{sec:gameImportance}). Second,  we apply a completeness safeguard that prevents the release of an analysis table unless the experimental grid is exactly balanced and the recorded payoffs for each game correctly recover the payoff scale assigned to its corresponding cell (i.e. a  combination of experimental conditions, such as model, language, persona, and payoff scale). Third, we include a strategy read-out that identifies which canonical memory-one rule each trajectory most closely resembles, while also reporting the provenance of that classification (Section \ref{sec:readout}).

The design is a complete factorial crossing of five language models, ten payoff scales, five prompt languages, four persona pairings and ten replicates. The unit of collection is the dyad, one ten-round game between two agents of the same model, contributing two agent-games, an agent-game being the ten decisions one agent takes in one game. Each of the 250 model-by-language-by-scale cells holds exactly 40 dyads, the four persona pairings by ten replicates; pooling the five languages gives 200 dyads and 400 agent-games in each of the 50 model-by-scale cells. The main-text corpus is 10{,}000 dyads, 20{,}000 agent-games and 200{,}000 decisions, and is the grid every main-text figure and table is computed on. Balance is enforced rather than assumed (appendix~A.1).

\subsection{The payoff-scaled prisoner's dilemma and its invariance}\label{sec:gameImportance}

Both agents receive the prisoner's-dilemma prompt of the FAIRGAME protocol~\cite{buscemi2025fairgame} in its no-communication, announced-horizon condition. The scenario casts the strategic interaction as a criminal interrogation, states the four payoff outcomes as penalties, names the two available actions Option~A and Option~B, carries the running history of prior moves and realised penalties, announces a fixed horizon of $T = 10$ rounds, and instructs each agent to minimise its cumulative penalty.

The row player's baseline payoff matrix for the Prisoner's Dilemma is
\begin{center}
\small
\begin{tabular}{c|cc}
 & Option~A & Option~B \\
\hline
Option~A & $(6, 6)$ & $(0, 10)$ \\
Option~B & $(10, 0)$ & $(2, 2)$
\end{tabular}
\end{center}
where the first and second entries in each cell denote the penalty payoffs to the row and column player, respectively. In standard game-theoretic notation, this matrix assigns the four canonical outcomes:
\[
T = 0, \quad R = 2, \quad P = 6, \quad S = 10,
\]
representing the temptation ($T$), reward ($R$), punishment ($P$), and sucker ($S$) penalty payoffs. The penalty ordering
\[
T < R < P < S \quad \text{with} \quad 2R < T + S
\]
is the exact isomorphic image of the familiar canonical Prisoner's Dilemma condition $T_{\text{util}} > R_{\text{util}} > P_{\text{util}} > S_{\text{util}}$ and $2R_{\text{util}} > T_{\text{util}} + S_{\text{util}}$ defined on utilities $u = -\text{penalty}$:
\[
T_{\text{util}} = 0 > R_{\text{util}} = -2 > P_{\text{util}} = -6 > S_{\text{util}} = -10.
\]
Because lower penalties correspond to higher utilities, Option~A strictly dominates Option~B in every column (facing Option~A, choosing A yields penalty 6 versus 10; facing Option~B, choosing A yields penalty 0 versus 2). Option~A is therefore the defecting action, while mutual Option~B yields penalty 2 versus penalty 6 for mutual Option~A, making Option~B the cooperative action; throughout this paper, cooperation denotes choosing Option~B.

The realised score-action relationships reproduce the intended dilemma structure in the logs: own cooperation correlates with own utility at Spearman $\rho = -0.015$, while the opponent's cooperation correlates at $\rho = +0.916$, and dyad welfare correlates with joint cooperation at $\rho = +0.985$. These checks validate the payoff mapping and action polarity, not model comprehension of the dilemma (appendix~A.1).

\begin{figure}[!htbp]
\centering
\includegraphics[width=\linewidth]{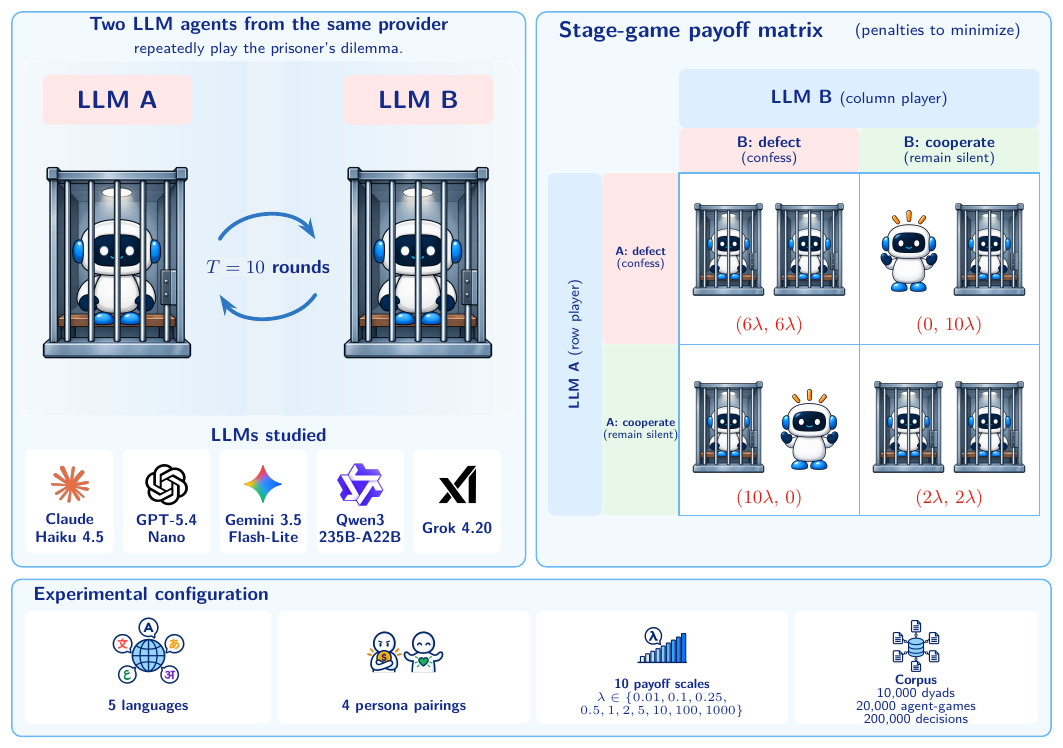}
\caption{\textbf{Overview of the repeated prisoner's-dilemma experiment.} Two language-model agents from the same provider repeatedly play the stage game across ten payoff scales, with penalties to minimise. The main-text corpus covers five widely adopted commercial models, five prompt languages, four persona pairings and 10{,}000 dyads (20{,}000 agent-games and 200{,}000 decisions).}
\label{fig:overview}
\end{figure}

To manipulate the stakes of the game without altering its strategic structure, best-reply correspondences, or Nash equilibria, we introduce a positive scalar parameter $\lam > 0$ and multiply all four baseline cells by $\lam$ (figure~\ref{fig:overview}). In our experiments, we evaluate ten values spanning five orders of magnitude:
\[
\lam \in \{0.01,\ 0.10,\ 0.25,\ 0.50,\ 1.0,\ 2.0,\ 5.0,\ 10.0,\ 100.0,\ 1000.0\},
\]
corresponding to attenuated ($\lam < 1.0$), baseline ($\lam = 1.0$), and amplified ($\lam > 1.0$) payoff regimes. For example, at the sub-unit scale $\lam = 0.10$, the row player's payoff matrix becomes
\begin{center}
\small
\begin{tabular}{c|cc}
 & Option~A & Option~B \\
\hline
Option~A & $(0.6, 0.6)$ & $(0, 1.0)$ \\
Option~B & $(1.0, 0)$ & $(0.2, 0.2)$
\end{tabular}
\end{center}
where every  payoff entry is at most 1.0. At the amplified scale $\lam = 10.0$, it becomes
\begin{center}
\small
\begin{tabular}{c|cc}
 & Option~A & Option~B \\
\hline
Option~A & $(60, 60)$ & $(0, 100)$ \\
Option~B & $(100, 0)$ & $(20, 20)$
\end{tabular}
\end{center}
with the ordering $T < R < P < S$ strictly preserved in all cases. The complete matrix for all ten scales is provided in appendix~A.3.

Under classical decision theory and game theory, multiplying utilities by $\lam > 0$ constitutes a positive affine transformation of a von Neumann-Morgenstern utility function~\cite{von2007theory,luce1957games}:
\[
u'(a) = \lam u(a) + c \quad (\lam > 0,\ c = 0).
\]
Such transformations preserve all preference orderings over lotteries, best replies, Nash equilibria, and deterministic replicator dynamics orbits. Consequently, the normative theoretical prediction for the effect of $\lam$ on rational agents is exactly zero. \rev{This is a result about the game. It predicts unchanged behaviour only if the stated penalties are the agent's whole utility and its choices do not depend on their numerical form; the criminal-interrogation narrative and the persona instruction do not scale with $\lam$ and may add considerations beyond the penalties. The null is therefore a prediction for an agent that meets these assumptions.} This construction isolates the effect of payoff magnitude from structural incentives: any systematic departure from invariance constitutes an empirical discovery about machine behaviour rather than a rational adaptation to changing game-theoretic payoffs.

\subsection{Models, prompt languages, personas and protocol}\label{sec:protocol}

Five widely adopted commercial models were run, each against itself; two different models were never paired. They were Claude Haiku 4.5, GPT-5.4 Nano, Gemini 3.5 Flash-Lite, Qwen3 235B-A22B and Grok 4.20. Their model identifiers are listed in appendix~A.2.

The prompt languages are English, French, Vietnamese, Chinese and Arabic, in templates byte-identical to the FAIRGAME resource files and checked against them before every collection run. Artefacts of those files are reproduced rather than repaired, the consequential one being that the Arabic and Chinese versions state the four cells as penalties but phrase the goal sentence as maximising a reward, so we treat the five languages as five distinct stimuli rather than five renderings of one (appendix~A.4). Each agent also receives one line naming its persona, in the language of the prompt: in English, \emph{You are cooperative.} or \emph{You are selfish.} The two are crossed in all four ordered pairings, ten replicates each. The opponent's persona is never disclosed, so a pairing reaches a player only through realised behaviour and not before round 2, and we treat it as a design factor and a regression covariate rather than as information the agent held.

Every game runs ten rounds with simultaneous moves: both agents see the same history, rounds 1 to $r-1$, and the round is scored once both have replied. Each game is repeated at all ten payoff scales with the same persona pairing and the same requested sampling seeds (appendix~A.2). Decisions are elicited as free text and parsed by the protocol's own matcher. A reply it cannot resolve is regenerated, up to two further attempts, and if all three fail the decision is recorded as Option~A, which under this framing is defection: the fallback can only push a measured cooperation rate down, never up, and it cannot manufacture the cooperative behaviour reported here. Each collection run asserts that its fallback rate is at most 0.02 (appendix~A.2). \rev{The cost of the collection is reported in appendix~A.2.}

\subsection{Strategy read-out}\label{sec:readout}

While the framework provides complete gameplay trajectories and descriptive metrics such as cooperation rates, these capture what agents do rather than the decision rule behind it. Following prior work on inferring canonical strategies from noisy repeated-game data~\cite{han2013intention,Han2012ALIFEjournal,diStefanoIntention2023}, we read every agent-game against the four canonical rules AllC, AllD, Tit-for-Tat and Win-Stay-Lose-Shift. The read-out reports which of the four a ten-round trajectory most \emph{resembles}. It is not an identification of the strategy a model is running, and no claim in this paper treats it as one.

All four rules are memory-one, so exact matching is a lookup with no learning in it, and it returns a \emph{set} rather than a label, because over ten rounds several rules routinely prescribe the same moves. The three provenances, the agent-games matched exactly by one rule (deduced), by several at once (ambiguous) and by none (unmatched), are recorded separately and never merged, and the same matcher yields a continuous quantity needing no labels at all, the number of rounds violating the nearest rule. Where the rules leave the answer open a single-layer LSTM~\cite{lstm} arbitrates. Trained on a synthetic corpus alone, which contains no language-model output, it reaches 98.5\% accuracy in distribution and 93.6\% on a held-out execution-noise condition (appendix~A.6). It ranks only within the set the rule base returns, so it can never overturn a deduction; it decides alone only where no rule fits, and its share of the labels is very unevenly spread across the four rules, which bounds how the composition may be read (Section \ref{sec:llm_strategies}).

\subsection{Statistical analysis}\label{sec:stats}

The cooperation rate of an agent-game is the share of its ten decisions that are Option~B; a cell mean averages the agent-games of a cell, and round-level analyses use the binary decision itself. Because the two agents of a dyad play each other and the ten rounds inside an agent-game are strongly autocorrelated, primary inferential intervals were computed by resampling whole dyads with replacement, 1{,}500 draws, percentile method~\cite{efron1993introduction}, and primary per-model and trend regressions clustered their standard errors on the dyad. \rev{Across scales, the ten dyads that share a model, a language and a game identifier, and so a persona pairing and the requested seeds, form a matched ladder, 200 per model. Appendix~D.6 repeats the headline inferences with the ladder as the unit; no conclusion changes.} The supplementary scale-by-language robustness model instead clusters on the documented language-by-game-ID blocks to preserve the matched ten-scale ladders. Descriptive figure intervals use whole-dyad resampling. Because the range over the ten scales is a maximum minus a minimum, and so positive by construction, its bootstrap interval quantifies the precision of the range and not its distance from zero; the permutation test supplies the test. Because the null is a theorem, the matching test destroys the association with the payoff scale and leaves the rest of the design as observed: the scale label is shuffled among whole dyads, the unit to which the scale was assigned, and the range of the ten scale means recomputed, 2{,}000 draws, so $0.0005$ is the smallest attainable $p$ value.

For each model we fitted a logistic regression of the round-level decision on the payoff scale as a ten-level factor, with prompt language, own persona and opponent persona as covariates, $\lam = 1$ as reference and standard errors clustered on the dyad, and tested the nine scale contrasts jointly by a Wald test (table~\ref{tab:effect}). The variance decomposition is a type-II analysis of variance on the 250 cell means of the five models reported here, with terms for model, language and payoff scale and their three two-way interactions and no three-way term, so the residual on 144 degrees of freedom is the omitted three-way interaction; the supplementary decomposition is in appendix~B.3. The persona result contrasts the two sub-unit scales with the remaining eight as a difference of differences with an interval from the same dyad bootstrap; where a trend across the whole grid is wanted instead, $\log_{10}\lam$ enters as a continuous predictor. Analyses were run in Python; the package versions and the seeds are recorded with the archived pipeline in appendix~E.

\section{Results}\label{sec:results}

The results below are computed on the 200{,}000 recorded decisions in the five-model main-text corpus.

\subsection{Cooperation varies under payoff rescaling }\label{sec:payoff_magnitude}

Under normative decision theory and classical game theory, multiplying every payoff in a normal-form matrix by a positive constant is an axiomatically neutral transformation. It preserves all preference rankings over lotteries, best-reply correspondences, and Nash equilibria: an agent playing the game should not alter its choices whether payoffs are denominated in cents, dollars, or thousands of pounds.

\begin{figure}[!htbp]
\centering
\includegraphics[width=\linewidth]{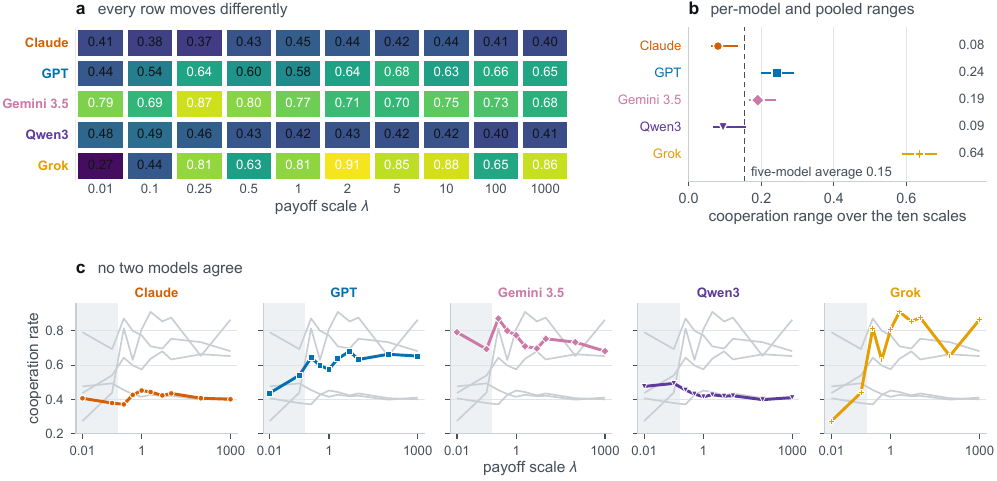}
\caption{\textbf{Cooperation rates as a function of the payoff scale across five models.} (a) Model-by-scale cooperation matrix: mean cooperation for each of the five models at each of the ten payoff scales $\lam \in [0.01, 1000]$, with the value shown in each box. (b) Within-model cooperation range (maximum minus minimum rate across the ten scales) compared against the pooled five-model range (0.154, vertical dashed line). \rev{The pooled range is narrower than the within-model range of three of the five models.} Confidence intervals are 95\% percentile intervals from 1{,}500 whole-dyad bootstrap resamples. (c) Small multiples per model: the focal model is drawn with its designated colour and marker, with the other four models shown behind in grey for context. Shaded bands mark the sub-unit regime ($\lam \leq 0.1$). Each model-by-scale cell pools 200 dyads (400 agent-games), 20{,}000 agent-games in all.}
\label{fig:landscape}
\end{figure}

Yet, across five orders of payoff magnitude, every language model evaluated violates this scale invariance (see figure~\ref{fig:landscape}a,c, table~\ref{tab:effect}). The shift in strategic behaviour is substantial: across the ten-scale ladder, cooperation varies markedly for every model, representing a clear departure from the scale-invariance null (see figure~\ref{fig:landscape}b and table~\ref{tab:effect}). What classical game theory treats as an inert change of units produces a fundamental reordering of multi-agent behaviour.

Notably, the models do not share a universal behavioural pattern: each possesses a distinct machine character that responds idiosyncratically to the scale of the stakes. Grok exhibits a \rev{strong behavioural} shift across payoff scales. At micro-stakes, it predominantly  defects; at standard stakes, however, it shifts towards high levels of cooperation, selecting the socially optimal action in the vast majority of interactions. GPT climbs sharply from a hesitant baseline before saturating at a stable cooperative tendency. Claude’s cooperation rate decreases  at intermediate sub-unit stakes before recovering at larger payoff scales. Gemini maintains an increased  cooperative tendency throughout, while Qwen becomes slightly less cooperative as the stakes rise (most of the \rev{change} occurs between $\lam = 0.1$ and $\lam = 1$). Because each model follows a distinct behavioural trajectory, machine cooperation cannot be characterised by a single universal behavioural pattern.

The largest departures are concentrated near the low-payoff end of the grid (table~\ref{tab:effect}). Fractional payoff multipliers often reduce cooperation compared with intermediate scales, whereas changes from unit stakes to large stakes have no consistent pattern across models. This result should be interpreted descriptively. The present design does not establish a discrete cognitive threshold, distinguish numerical magnitude from every feature of notation, or identify the mechanism behind the response.

\begin{table}[t]
\centering
\caption{\textbf{Cooperation over the ten payoff scales, by model.} Minimum, maximum and range, with a 95\% confidence interval on the range from a bootstrap over whole dyads and a permutation test on the payoff-scale label ($0.0005$ is the floor of 2{,}000 draws). The last two columns test the nine scale contrasts jointly in a per-model logistic regression of the round-level decision, clustered on the dyad. Each entry pools 200 dyads per scale; ranges are computed before rounding. Full specification in the supplementary material.}
\label{tab:effect}
\small
\begin{tabular}{lccccccc}
\toprule
& \multicolumn{2}{c}{cooperation} & & & & \multicolumn{2}{c}{Wald test} \\
\cmidrule(lr){2-3} \cmidrule(lr){7-8}
Model & min & max & range & 95\% CI & $p_{\text{perm}}$ & $\chi^2_9$ & $p$ \\
\midrule
Claude Haiku 4.5 & 0.371 & 0.452 & 0.081 & [0.062,\,0.135] & $0.0025$ & 32.3 & $<0.001$ \\
GPT-5.4 Nano & 0.438 & 0.680 & 0.243 & [0.200,\,0.290] & $0.0005$ & 164.1 & $<0.001$ \\
Gemini 3.5 Flash-Lite & 0.682 & 0.872 & 0.190 & [0.165,\,0.240] & $0.0005$ & 129.0 & $<0.001$ \\
Qwen3 235B-A22B & 0.399 & 0.494 & 0.095 & [0.068,\,0.158] & $0.0005$ & 96.9 & $<0.001$ \\
Grok 4.20 & 0.274 & 0.910 & 0.636 & [0.588,\,0.686] & $0.0005$ & 1227.3 & $<0.001$ \\
\bottomrule
\end{tabular}
\end{table}

\begin{table}[t]
\centering
\caption{\textbf{Persona effect below and above the sub-unit boundary, by model.} Cooperation under a cooperative persona minus cooperation under a selfish one, at the two scales where every shown payoff is at most 1 ($\lam \leq 0.1$) and at the eight where some payoff exceeds it. Intervals are 95\% confidence intervals from the dyad bootstrap. A sign flip needs both regime intervals to exclude zero on opposite sides, so a supra-unit interval straddling zero means the effect is abolished rather than reversed.}
\label{tab:persona}
\footnotesize
\setlength{\tabcolsep}{3pt}
\begin{tabular}{lccccc}
\toprule
& \multicolumn{2}{c}{persona effect, with 95\% CI} & & & \\
\cmidrule(lr){2-3}
Model & $\lam \leq 0.1$ & $\lam \geq 0.25$ & shift & 95\% CI & sign flip \\
\midrule
Claude Haiku 4.5 & $-0.222$ $[-0.258,\,-0.186]$ & $+0.207$ $[+0.189,\,+0.226]$ & 0.429 & [0.389,\,0.471] & yes \\
GPT-5.4 Nano & $-0.135$ $[-0.179,\,-0.091]$ & $+0.018$ $[-0.001,\,+0.037]$ & 0.153 & [0.105,\,0.200] & no \\
Gemini 3.5 Flash-Lite & $-0.456$ $[-0.507,\,-0.406]$ & $+0.224$ $[+0.207,\,+0.242]$ & 0.680 & [0.625,\,0.736] & yes \\
Qwen3 235B-A22B & $-0.911$ $[-0.931,\,-0.893]$ & $-0.679$ $[-0.694,\,-0.664]$ & 0.233 & [0.210,\,0.258] & no \\
Grok 4.20 & $-0.441$ $[-0.485,\,-0.396]$ & $-0.303$ $[-0.326,\,-0.279]$ & 0.138 & [0.090,\,0.185] & no \\
\bottomrule
\end{tabular}
\end{table}

\begin{table}[t]
\centering
\caption{\textbf{Rule agreement and rule distance against the payoff scale, by model.} Left, the share of agent-games matched exactly by one canonical rule, below and above the sub-unit boundary, with the coefficient of a logistic regression of that indicator on $\log_{10}\lam$; right, the mean number of rounds violating the nearest canonical rule at the smallest and largest scale, with the same trend by least squares. Both $\beta$ are per decade of $\lam$ and their signs differ between models. Both cluster on the dyad, and neither block uses the learned classifier.}
\label{tab:strategy}
\small
\setlength{\tabcolsep}{4pt}
\begin{tabular}{lcccccccc}
\toprule
& \multicolumn{4}{c}{exactly one rule fits (\%)}
& \multicolumn{4}{c}{rounds violating the nearest rule} \\
\cmidrule(lr){2-5} \cmidrule(lr){6-9}
Model & $\lam \leq 0.1$ & $\lam \geq 0.25$ & $\beta$ & $p$
      & $\lam{=}0.01$ & $\lam{=}10^3$ & $\beta$ & $p$ \\
\midrule
Claude Haiku 4.5 & 17.9 & 3.5 & $-0.545$ & $<0.001$ & 1.70 & 2.02 & $+0.073$ & $<0.001$ \\
GPT-5.4 Nano & 9.8 & 7.6 & $-0.038$ & $0.441$ & 2.02 & 2.19 & $+0.038$ & $0.052$ \\
Gemini 3.5 Flash-Lite & 32.2 & 15.6 & $-0.242$ & $<0.001$ & 0.48 & 1.36 & $+0.194$ & $<0.001$ \\
Qwen3 235B-A22B & 72.1 & 53.6 & $-0.267$ & $<0.001$ & 0.27 & 1.14 & $+0.202$ & $<0.001$ \\
Grok 4.20 & 44.6 & 17.1 & $-0.301$ & $<0.001$ & 0.92 & 0.37 & $-0.118$ & $<0.001$ \\
\bottomrule
\end{tabular}
\end{table}

\subsection{Effect sizes vary across prompt languages}\label{sec:language}

\begin{figure}[!htbp]
\centering
\includegraphics[width=\linewidth]{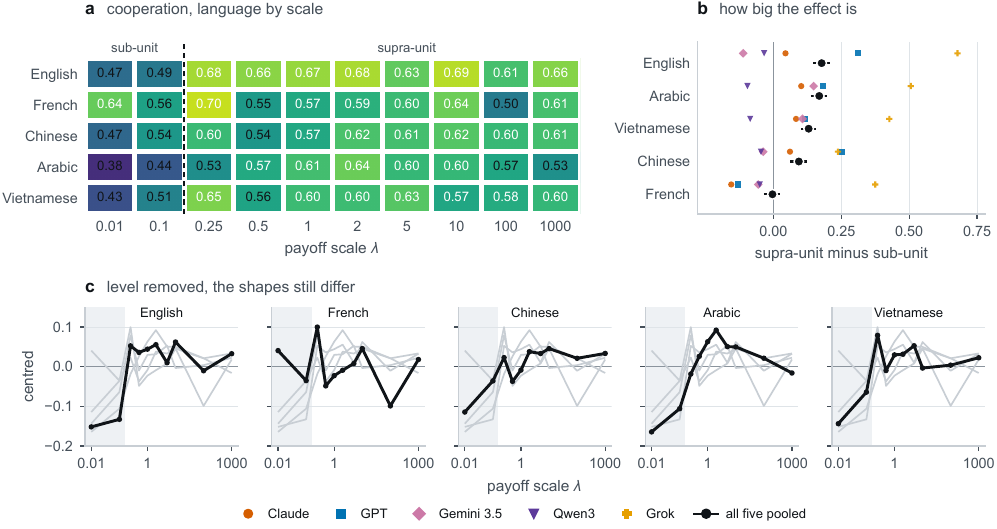}
\caption{\textbf{Cross-lingual scale sensitivity.} (a) Cooperation pooled across the five main-text models for each prompt language and payoff scale; the dashed boundary separates the sub-unit and supra-unit regimes. (b) Language-specific difference in cooperation between supra-unit and sub-unit regimes; intervals show the pooled estimate and small markers the five models separately. (c) Cooperation curves after subtracting each language's mean, showing that differences in scale-response shape remain after level differences are removed.}
\label{fig:language}
\end{figure}

Prompt language is not a neutral transmission channel for the game. Both the baseline level of cooperation and its sensitivity to the stakes diverge substantially across prompt languages. Figure~\ref{fig:language} shows the pooled language profiles, while the model-by-language levels and ranges appear in table~B.3.

We note a caveat related to the comparison of levels rather than the scale effect itself. The Arabic and Chinese templates differ from the other prompt stimuli in their objective wording. They state the four cells as penalties while phrasing the goal sentence as maximising a reward (Section \ref{sec:protocol}). Because that wording is fixed within each language, the within-language comparisons still establish sensitivity to payoff rescaling for the prompt as actually presented. However, differences in the magnitude or shape of scale sensitivity across languages cannot be attributed to language alone, because language is partially confounded with template wording. We therefore interpret the scale-by-language interaction as a prompt-stimulus interaction, not as evidence for a linguistic or cultural mechanism.

The central finding is that scale sensitivity itself depends on the prompt language. The max-minus-min range over the ten scales differs between languages by a factor of about two within every model, and by an order of magnitude across model-language cells (table~B.3). Figure~\ref{fig:language}b provides the complementary supra-unit minus sub-unit contrast, and figure~\ref{fig:language}c shows that language differences remain after removing their mean levels. Across the main-text corpus, the observed sensitivity ranges from near-invariance to sweeping behavioural transformations spanning almost the entire probability scale.

Crucially, these linguistic biases do not point in a uniform direction across models: the most cooperative language differs from model to model, whereas English is the most scale-sensitive language for four of the five models. When multilingual evaluations pool across models, these opposing model-specific effects are attenuated and can obscure substantial model-by-language heterogeneity. In reality, machine behaviour depends on the linguistic stimulus used in the prompt. Within this design, scale sensitivity is a joint property of the model and prompt language rather than a model-independent invariant. It does not identify a cultural mechanism.

\subsection{Model heterogeneity and the effects of aggregation}\label{sec:heterogeneity}

Averaging across models conceals the real phenomenon through mutual cancellation. The pooled response curve severely attenuates the true magnitude of scale sensitivity, shrinking the typical within-model movement substantially (figure~\ref{fig:landscape}b). Because six of the ten pairwise correlations between model curves are negative (table~B.11), curves that move in opposite directions flatten each other out when summed. A pooled curve describes an artificial average agent that no real model resembles.

Underlying this cancellation is a fundamental divide in policy geometry: cooperation in these models is not a single Gaussian distribution, but a mixture over distinct policy regimes (appendix~B.6). Across the corpus, unconditional boundary play is strongly overrepresented, with extreme atoms carrying far more weight than expected under uniform play. Crucially, the commitment to boundary play varies dramatically across models: Claude operates as an interior, mixed-policy player that rarely locks into extreme  behavioural patterns, whereas Qwen and Grok function as pure-policy switchers, oscillating between unconditional defection and unconditional cooperation. Pooling across models conflates how often a model chooses a pure strategy with which pure strategy it selects.

A formal variance decomposition confirms that model individuality dominates the game (table~B.4, appendix~B.3). Model identity and the model-by-language interaction account for the majority of the variance across cells. While the main effect of payoff scale is statistically detectable, its interaction with model identity carries three times as much variance, demonstrating that the behavioural impact of stakes is model-specific rather than shared universally across model endpoints.

In the five-model cell-mean variance decomposition, the shared payoff-scale main effect is largely carried by Grok. Removing Grok renders that particular main-effect test non-significant, even though every remaining model continues to move (appendix~B.3). A complementary dyad-level leave-one-model-out test asks a different question, whether any pooled scale association remains. Removing any one model, from the five reported here or from all six, leaves a detectable pooled association, including the fits without Grok (appendix~D.6). The two analyses are therefore complementary: one tests a shared main effect after accounting for heterogeneity, whereas the other tests any residual pooled association.

\subsection{Payoff scale modulates persona-induced cooperative behaviour}\label{sec:persona}

\begin{figure}[!htbp]
\centering
\includegraphics[width=\linewidth]{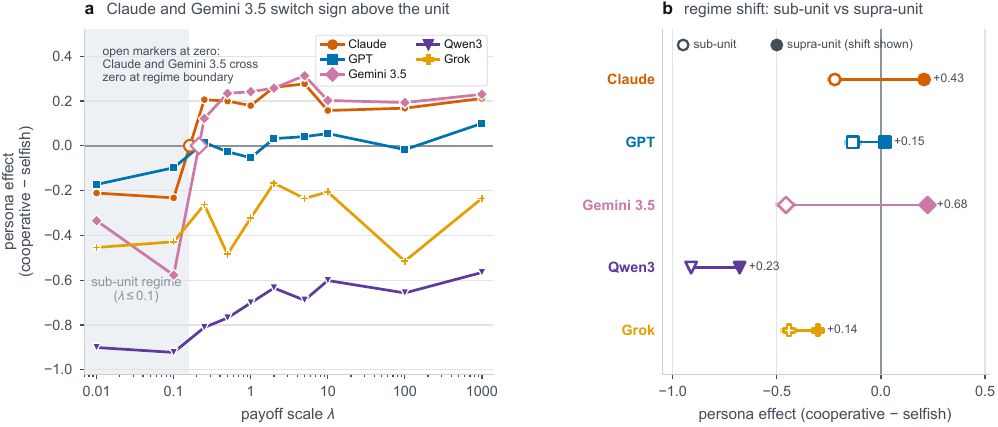}
\caption{\textbf{Persona effect and sub-unit inversion against the payoff scale.} The persona effect is the cooperation rate under the cooperative persona minus that under the selfish persona. (a) Difference in cooperation rate across scales $\lam$, with the sub-unit regime ($\lam \le 0.1$) shaded where all five models cooperate more under the selfish prompt. (b) Regime means comparing sub-unit ($\lam \le 0.1$, open circles) with supra-unit ($\lam \ge 0.25$, filled circles) regimes (table~\ref{tab:persona}).}
\label{fig:persona}
\end{figure}

System prompts designed to steer agent behaviour do not operate independently of game stakes. The behavioural difference between cooperative and selfish persona instructions changes dramatically across the payoff ladder (figure~\ref{fig:persona}, table~\ref{tab:persona}). This modulation differs between the two sub-unit scales and the eight larger scales, where shown payoffs include values above one (Section \ref{sec:gameImportance}). Across all five models, the persona effect changes across these two regimes.

In several models, the persona instruction reverses its behavioural effect  below the unit threshold. In Gemini and Claude, the persona effect reverses sign: under fractional stakes, both models exhibit higher cooperation when explicitly instructed to behave selfishly than when instructed to behave cooperatively. In GPT, the selfish persona also raises cooperation at fractional stakes, but above them the persona effect disappears instead of reversing, while the remaining models maintain their pattern. This result is a boundary-aligned descriptive effect. It does not identify why the prompt instruction changes direction at these scales.

The main grid does not separate numerical magnitude from all aspects of numeral formatting. We therefore treat this regime split as descriptive and do not attribute it to a specific representational mechanism.

A separate class of models inverts the persona instruction across the entire ladder (table~\ref{tab:persona}). Qwen and Grok cooperate substantially more under the selfish persona than under the cooperative persona at every evaluated scale. These models show an opposite response to the persona wording. The data do not establish why they do so. This result underscores that role prompts cannot be assumed to steer multi-agent interactions in the intended direction without empirical verification across payoff contexts (appendix~B.4).

\subsection{ Payoff scale alters strategic composition, not merely cooperation rates}\label{sec:llm_strategies}

Two models displaying identical cooperation rates can be executing fundamentally different behavioural strategies/rules. To understand the behavioural rules underlying raw cooperation rates, we read every agent trajectory against the four canonical memory-one strategies of evolutionary game theory (Section \ref{sec:readout}). What this read-out captures is behavioural resemblance rather than exact algorithmic identity: across the corpus, the majority of games match no textbook rule cleanly (table~\ref{tab:strategy}, appendix~B.5).

As payoff stakes increase, play moves farther from the four canonical rules in most models. Across the scale ladder, the proportion of games adhering strictly to a single canonical rule declines substantially, while unmatched play accumulates (table~\ref{tab:strategy}). This loss of canonical-rule conformity is pronounced in Claude, Grok, Qwen, and Gemini, while GPT maintains steady adherence. The read-out does not establish what rule generates unmatched play; it only shows that it is farther from the four canonical rules.

A label-free continuous metric confirms this behavioural drift: the number of rounds departing from the nearest canonical rule increases significantly with scale across most models (table~\ref{tab:strategy}). As stakes multiply, trajectories depart more often from the textbook templates. Only Grok reverses this trend, moving towards extreme pure-policy play at high stakes.

\begin{figure}[!htbp]
\centering
\includegraphics[width=\linewidth]{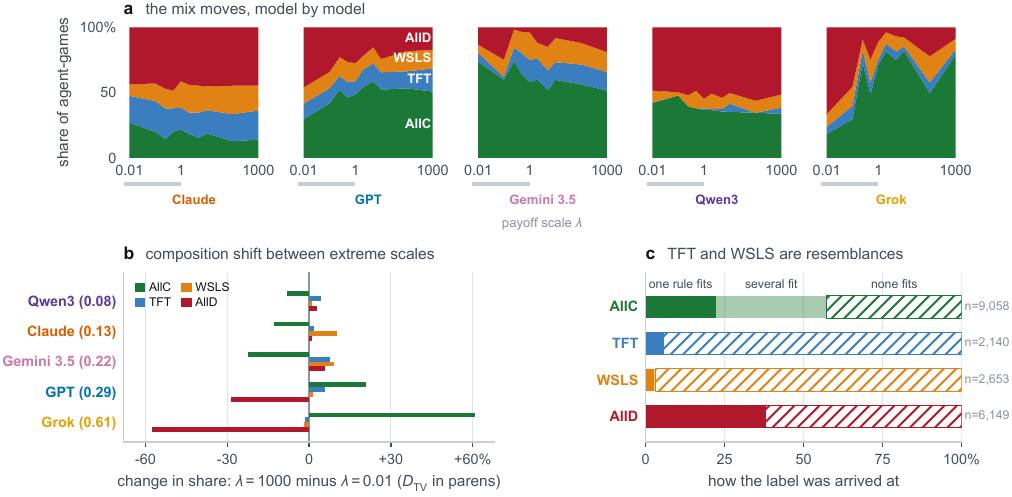}
\caption{\textbf{Strategy attribution across payoff scales, composition shift, and provenance breakdown.} (a) Stacked area share of canonical memory-one strategy attributions (AllC, AllD, TFT, WSLS) across the ten payoff scales for each of the five models. (b) Net strategy share changes between extreme scales ($\lam = 10^{3}$ minus $\lam = 0.01$) across the four canonical rules with total variation distance $D_{\mathrm{TV}}$ in parentheses. (c) Strategy read-out provenance: exactly deduced (single canonical match), ambiguous (multiple matching rules), and unmatched (LSTM attribution on trajectories consistent with no standard rule). Over the corpus, $94.6\%$ of Tit-for-Tat and $97.1\%$ of Win-Stay-Lose-Shift labels are attributions on unmatched play, establishing that labels indicate resemblance rather than faithful execution.}
\label{fig:mix}
\end{figure}

The qualitative composition of strategies also shifts dynamically, driving divergent realignments across models (figure~\ref{fig:mix}). In GPT, expanding stakes prompts a steady migration from unconditional defection toward unconditional cooperation. In Gemini, the opposite shift occurs: unconditional cooperation yields ground to more Tit-for-Tat-like and Win-Stay-Lose-Shift-like attributions. Grok undergoes an outright reversal, swapping rampant defection at micro-stakes for overwhelming cooperation at high stakes.

Critically, these strategy shifts reflect qualitative resemblances rather than exact implementations \rev{of the canonical strategies}. Provenance auditing demonstrates that while unconditional cooperation and defection are often matched exactly, nearly all attributions of Tit-for-Tat and Win-Stay-Lose-Shift represent best-fit approximations to noisy, unmatched play (figure~\ref{fig:mix}c, appendix~B.5). When models appear to increase their Tit-for-Tat share, they are not adopting the classic reciprocal algorithm; they are generating complex, imperfect trajectories whose overall statistical contours happen to lean in that direction.

\subsection{Where payoff scale  acts: the evolutionary baseline}\label{sec:egt_pointer}

While payoff scale is theoretically inert in normal-form games, evolutionary game theory provides a foundational framework where stakes legitimately matter, offering a rigorous theoretical benchmark for language model behaviour.

In deterministic replicator dynamics, scale is completely neutral. Multiplying the payoff matrix by a positive constant simply scales the velocity of the vector field: orbits follow identical geometric trajectories and all rest points remain invariant (appendix~C.2, figure~C.1).

\begin{figure}[!htbp]
\centering
\includegraphics[width=\linewidth]{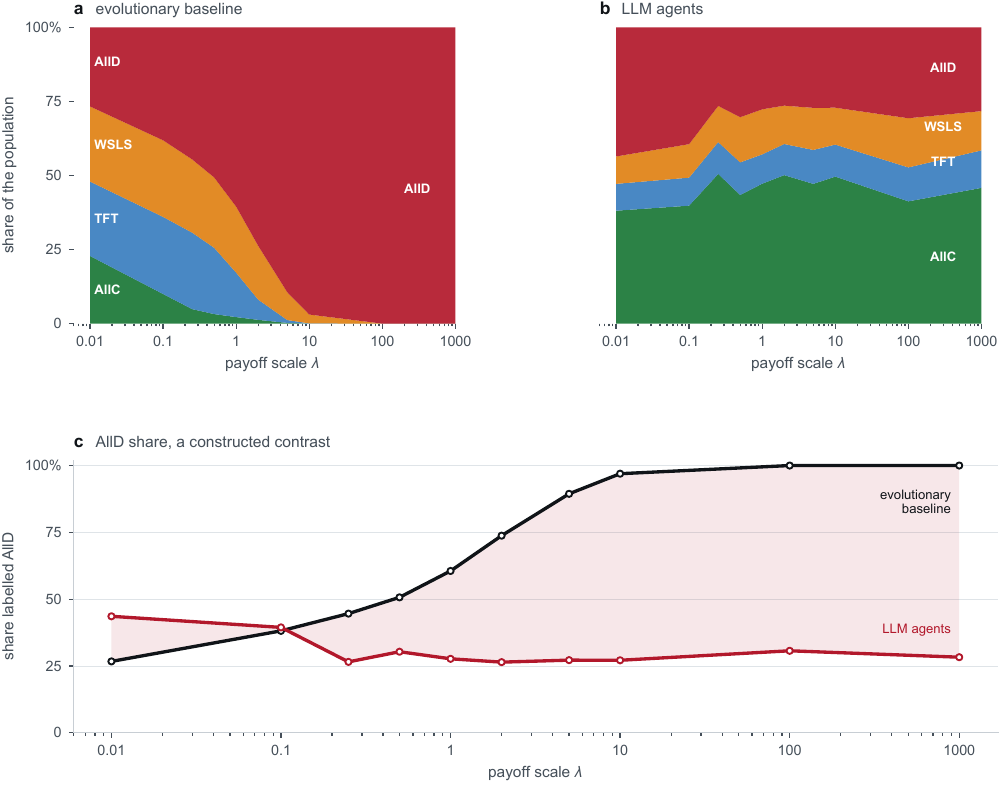}
\caption{\textbf{The evolutionary baseline and the corpus move in opposite directions.} (a) The baseline's stationary distribution over the four canonical rules across the ten payoff scales, drawn as a stacked band. (b) The corpus's strategy mix on the same axis, drawn identically. (c) The AllD share alone, both accounts on one axis, with the gap shaded. \rev{Panel (c) is a constructed contrast between two different objects, selection among rules in the baseline and fixed language-model agents in the corpus, so the gap is descriptive and not a test of one against the other.} Baseline at $Z = 100$, $\beta = 0.1$, $\epsilon = 0.05$, ten rounds; the corpus pooled over five models, five languages and four persona pairings.}
\label{fig:egtllm}
\end{figure}

In finite populations with stochastic imitation \citep{nowak2004emergence}, however, payoff scale acts directly through the selection intensity. Under pairwise Fermi imitation, scaling payoffs amplifies the competitive advantage of defection over cooperation (appendix~C).  \rev{With $\beta$ fixed, payoffs enter the Fermi rule only through $\beta\lam$, so raising $\lam$ is equivalent to raising the intensity of selection.} At micro-stakes, selection is weak and all four canonical rules coexist near genetic drift. As stakes expand, selection pressure intensifies: unconditional defectors strongly invade cooperative strategies until AllD achieves near-total fixation (figure~\ref{fig:egtllm}a). Combining evolutionary dynamics with language-model evaluations provides a principled lens for multi-agent governance~\cite{balabanova2025media,buscemi2025llms,xie2026evolutionary,han2026social}.

At the pooled five-model level, the AllD label share moves in the opposite direction to the finite-population evolutionary baseline as $\lam$ increases (figure~\ref{fig:egtllm}). This pooled reversal is driven primarily by GPT and Grok; per-model label trajectories remain heterogeneous. As payoffs grow into standard integer ranges, the pooled language-model corpus becomes more cooperative, whereas the evolutionary reference becomes more dominated by defection (figure~\ref{fig:egtllm}b,c).

This direction is stable across the eighty evolutionary parameter settings reported in the appendix: the stationary AllD share is higher at the largest than at the smallest scale in every setting, and the rise is strictly monotonic in 61 of 80 settings. None reproduces the direction of the pooled empirical trajectory. \rev{From $\lam = 0.25$ upward the corpus resembles AllD in 0.26 to 0.31 of agent-games against 0.45 to 1.00 in the baseline, the direction a prosocial preference would produce, although the prompt effects discussed in Section~\ref{sec:discussion} could produce it too.} We therefore treat the comparison as a descriptive reference gap rather than as evidence for a particular mechanism. Evolutionary dynamics describe selection among rules, whereas the corpus describes fixed language-model agents.

\rev{Figure~\ref{fig:egtllm}c is therefore a constructed contrast:} the gap is a descriptive difference between selection among rules and fixed language-model agents. The language conditions also vary the stimulus, so this comparison does not identify a cultural or cognitive mechanism.

\subsection{Agents condition more strongly on their own past actions}\label{sec:conditioning}

The evolutionary model rests on an explicit behavioural mechanism that can be evaluated directly in the data. In the classical account, cooperation survives invasion by defectors through \emph{reciprocity}: conditioning one's choice on what the partner just did, retaliating against defection and rewarding cooperation. Tit-for-Tat is the prototypical exemplar of this logic.

\begin{figure}[!htbp]
\centering
\includegraphics[width=\linewidth]{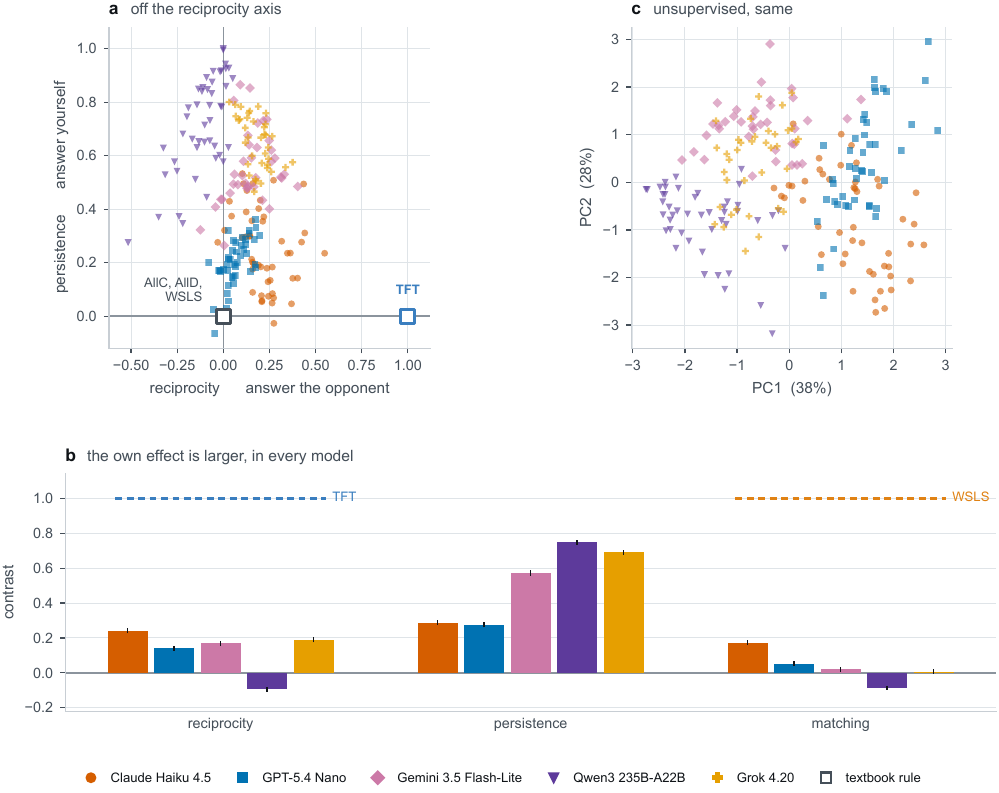}
\caption{\textbf{Own-action conditioning exceeds partner-action conditioning at the model level.} (a) Cells in the reciprocity by persistence plane, with the canonical rules at their exact coordinates; AllC, AllD and \rev{Win-Stay-Lose-Shift} all have both main effects zero and so coincide at the origin. (b) The three contrasts per model, against the value each canonical rule would give, with 95\% percentile intervals from whole-dyad bootstrap resampling. (c) The first two principal components of the raw five-vector, which is given no knowledge of the contrasts and recovers the same geometry. Panels (a) and (c) use model-by-scale-by-language cells; a predecessor observed fewer than thirty times yields no estimate and the cell is dropped, leaving 212 of 250.}
\label{fig:strategyspace}
\end{figure}

Because the previous round consists of an action pair (own action, partner action), the four conditional cooperation probabilities decompose cleanly into a base level and three orthogonal contrasts. \emph{Reciprocity} represents the partner main effect: the difference in cooperation after a partner cooperates versus after a partner defects. \emph{Persistence} represents the own main effect: the tendency to repeat one's own previous action. \emph{Matching} represents the interaction effect characteristic of Win-Stay-Lose-Shift. Textbook strategies occupy precise locations in this geometric space: Tit-for-Tat lies entirely on the reciprocity axis, Win-Stay-Lose-Shift on the matching axis, and unconditional rules at the origin (figure~\ref{fig:strategyspace}a).

Decomposing agent play reveals that own-action conditioning exceeds partner conditioning at the model level (figure~\ref{fig:strategyspace}). Across all five models, the persistence contrast is larger than the reciprocity contrast, although the margin is modest for Claude and substantially larger for the other models. Reciprocity remains positive in four models and is negative in Qwen. An unsupervised principal component analysis on raw transition vectors independently recovers this identical geometry without any prior knowledge of the contrasts (figure~\ref{fig:strategyspace}c).

The contrast estimates are precise at the dyad level. Whole-dyad bootstrap intervals for reciprocity are positive for Claude, GPT, Gemini~3.5, and Grok, and negative for Qwen; the corresponding persistence intervals are positive for all five models (table~D.2). Thus the headline comparison is one of relative strength, not a claim that partner information is absent.

This finding narrows one candidate explanation for the divergence from the evolutionary baseline. In these data, agents condition more strongly on their own preceding moves than on their partner's behaviour. The contrast does not show that partner information is absent, nor does it identify the causal basis of cooperation.

A technical caution governs this decomposition. The naive difference between disagreement states, $p_{CD} - p_{DC}$, is frequently mistaken for a reciprocity metric in the literature. In reality, it equals persistence minus reciprocity identically, confounding the two main effects and falsely indicating anti-reciprocal behaviour where the true partner effect is positive.

\section{Concluding remarks and discussion}\label{sec:discussion}

Positive scaling of payoffs is theoretically behaviour-preserving in the (repeated) prisoner's dilemma: \rev{it} does not alter the strategic structure of the game and therefore implies no change in predicted outcomes. Against this, our results show systematic departures from scale invariance across all five language models examined. These departures are not reducible to a homogeneous shift in baseline cooperation. Their magnitude varies across prompt languages, their direction and shape differ across models, and they are substantially attenuated by aggregation. The resulting aggregated trajectory is therefore not an adequate representation of any individual model's behavioural response. Relative to an evolutionary game baseline computed on the same payoff matrices, the aggregate corpus exhibits a decline in AllD-like behaviour as payoffs increase, contrary to the baseline prediction.

The observed violation of scale invariance has important implications for the interpretation of cooperation benchmarks. A cooperation rate can no longer be regarded as a scale-independent property of a model, but must instead be understood as contingent on how the interaction is represented. This point resonates with the machine behaviour perspective, which argues that autonomous systems should be studied empirically, using the tools of the behavioural sciences, because their behaviour cannot be inferred from their construction alone~\cite{rahwan2019machine}. Yet behavioural experiments face a familiar difficulty: most manipulations do not imply a precise predicted effect, so any observed change must be interpreted relative to a comparison condition chosen by the experimenter. The invariance classes of decision theory provide an unusual exception. They specify manipulations for which the predicted behavioural effect is exactly zero. A violation of such an invariance is therefore especially informative: it does not merely show that behaviour changes under a different experimental condition, but that it changes under a transformation that should be behaviourally inert.

This point has consequences for several strands of machine-behaviour research touched by this study. Persona prompting is the clearest case in which the experimenter addresses the system in terms of its social role; our results show that whether such an instruction takes hold can depend on the numerical units in which the payoff matrix is presented. The implication is equally direct for models of mixed human--machine populations, which treat the cooperativeness of artificial agents as a parameter and ask where the population settles~\cite{santos2018social,han2013good,perc2017statistical,paiva2018engineering}. A cooperation rate partly determined by the experimenter's choice of units is not the scale-independent quantity such models require.

The methodological implication of our findings is significant. Cooperation rates elicited from language models are often treated as properties of the model under evaluation, conditional on a prompt and a game form. The present evidence indicates that they are instead joint properties of the model, the prompt language, the persona instruction, and the numerical scale at which the payoff matrix is expressed. The last of these is especially important because it is usually treated as theoretically irrelevant. In standard game-theoretic analysis, multiplying every payoff-matrix entry by a positive constant changes units but not incentives. For the models studied here, however, this formally irrelevant choice systematically changes observed behaviour. Thus, the payoff scale cannot be regarded as a neutral implementation detail in empirical studies of machine cooperation.

This conclusion is strengthened by the structure of the heterogeneity. The observed scale effects are not merely noisy fluctuations around an invariant response. They differ by model, interact with prompt language, moderate the behavioural effect of persona instructions, and alter the composition of inferred strategic patterns. These features are difficult to reconcile with sampling variability alone. They suggest that the models are sensitive to aspects of the numerical representation of the game that are not part of its strategic equivalence class. However, because the design manipulates payoff scale rather than the internal representation of number, notation, or stakes, the results should be interpreted as evidence of invariance violation rather than as evidence for any particular psychological or computational mechanism.

We also qualify the interpretation of persona effects. Persona instructions are often treated as relatively direct interventions on behavioural disposition, such as inducing more cooperative or more selfish play. The present results show that their effect may depend on the payoff scale at which the interaction is framed. In several LLMs, persona instructions reverse their behavioural effect below the unit threshold: under fractional stakes, models instructed to behave selfishly may cooperate more than those instructed to behave cooperatively. This pattern does not imply that the models possess stable preferences that are being inverted by small payoffs. Rather, it shows that the behavioural consequence of a verbal instruction is conditional on the numerical context in which that instruction is applied. Persona effects are therefore not separable from payoff representation.

The strategy-composition analysis extends the same conclusion beyond marginal cooperation rates. Payoff scale changes not only how often cooperation occurs, but also the kinds of action sequences that the models generate. These shifts indicate qualitative resemblances to canonical repeated-game strategies rather than exact implementations of them. The distinction is important. A ten-round trajectory cannot establish that a model is executing Tit-for-Tat, Win-Stay-Lose-Shift, Grim Trigger, or any other  behavioural strategy. What can be inferred is that some trajectories are more consistent with the behavioural signatures of such strategies than others, and that the prevalence of these signatures changes with payoff scale. Scale sensitivity therefore operates at the level of sequential structure as well as aggregate cooperation.

Prior studies have examined how cooperation changes under moral framing, pre-play communication, institutional mediation, or other contextual interventions \cite{buscemi2025strategic,lu2024llms,mao2025alympics,xie2026evolutionary}. Those manipulations can  alter the game as interpreted by the agent, because they change the semantic environment in which the decision is made. Payoff rescaling is different: under the standard formal model, it preserves the game. The fact that behaviour nevertheless changes indicates that language models may not respect the invariance classes assumed by game-theoretic representations. More broadly, if model behaviour is governed by a utility-like object that combines numerical payoffs with linguistic or representational features, then scale invariance will generally fail unless the numerical and non-numerical components transform together. Any theoretical account of machine play in strategic settings must therefore specify which transformations it regards as behaviourally equivalent.

\rev{These departures do not resemble human stake effects. These effects in human studies are generally modest, tend to run in  a  common direction within a game, and  are most evident when stakes are large relative to participants’ income ~\cite{camerer1999effects,kocher2008does,engel2011dictator,larney2019stake,cameron1999raising,andersen2011stakes,holt2002risk}. Here, however, changing only the numerical scale of the payoffs, without changing the strategic incentives, shifts cooperation by as much as 0.636, with effects in opposite directions across models and most pronounced at or below $\lam = 1$. Four explanations, not mutually exclusive, fit parts of the evidence: prompt text that does not scale with $\lam$, such as the narrative and the persona, consistent with the scale-dependent persona effects; the representation of numbers, since sub-unit payoffs are rare and tokenised differently, consistent with the change concentrating at or below $\lam = 1$; provider-specific post-training, consistent with Qwen and Grok inverting the persona at every scale; and a distorted human stake response learned from text. Appendix~F gives the evidence for each and a design that would test it. The present design cannot separate them, so the results identify a robust behavioural phenomenon but not its causal source.}

Several limitations follow from this design. First, the study examines five models, one repeated game, one family of prompt templates, and ten payoff scales. The results therefore support claims about the presence and structure of scale sensitivity in this corpus, not universal claims about all language models or all strategic environments. Second, the language comparison is bounded by the prompt templates used. Two templates describe payoffs as penalties while retaining a reward-maximization objective. Because this wording is fixed within each language across payoff scales, it does not undermine the within-language rescaling comparison. It may, however, interact with numerical magnitude and therefore limits any attribution of cross-language differences to language alone.
Third, the strategy analysis is necessarily approximate. Strategies are latent decision rules, whereas the data consist of finite action sequences. Ten rounds provide limited information for distinguishing among memory-one rules, especially when execution noise is present. The classifier was validated up to 20\% execution noise, corresponding to two deviations in a ten-round sequence, and some models operate near that boundary on average. The resulting labels should therefore be read as descriptive summaries of trajectory resemblance, not as evidence that the models implement the corresponding strategies. Longer horizons, richer strategy classes, and mixture-based estimators would be needed to make stronger claims about strategic implementation \cite{inferenceStrategies,hilbe2013memory,han2013intention}. 

\rev{Fourth, our design does not include a condition without numerical payoffs. A verbal baseline describing each outcome only by its rank would keep the ordinal structure while removing every magnitude, and would show whether numerical payoffs themselves change behaviour (appendix~F.3).}

Overall, our study shows that payoff scale, although formally irrelevant under standard game-theoretic invariance, systematically affects machine cooperation. The effect is model-specific, language-dependent, and consequential for both cooperation rates and inferred strategic composition. These findings call for caution in interpreting single-scale cooperation benchmarks and suggest that scale sensitivity should be reported as a standard robustness measure in studies of language models in games. More generally, they indicate that empirical work on machine strategic behaviour cannot assume that formal equivalences in the game are automatically respected by the agents that are prompted to play it.

\section*{Ethics statement}
This work did not require ethical approval from a human subject or animal welfare committee.

\section*{Data accessibility}
The game transcripts, prompt templates, trained strategy classifier and analysis code supporting this study are \rev{deposited at Zenodo, \url{https://doi.org/10.5281/zenodo.23627009}~\cite{han2026data}. The deposit holds the per-game tables of all 12{,}000 dyads, the prompts in all five languages, the raw collection records, the game code, the classifier and the analysis pipeline that regenerates every table and figure.}

\section*{Declaration of AI use}
Large language models are the object of study in this article. Generative AI was used only as an assistant to identify and fix bugs in code and to edit English grammar.

\section*{Author contributions}
T.A.H.: conceptualisation, methodology, supervision, project administration, writing (original draft), writing (review and editing). T.-K.H.: conceptualisation, software, formal analysis, visualisation, writing (original draft). D.-S.D.-M.: software, data curation, formal analysis, writing (original draft). T.-B.C., P.-H.L., H.-D.N., P.-Q.N.-L., M.-L.N.-V., H.-P.P., P.-H.P., T.-K.T., C.-N.T., H.T. and G.-T.T.-L.: investigation, data curation, validation, writing (review and editing). A.B.: methodology, resources, writing (review and editing). L.H.T.: conceptualisation,  supervision,  writing (review and editing). T.A.H., T.-K.H. and D.-S.D.-M. contributed equally to this work. All authors gave final approval for publication and agree to be held accountable for the work performed therein.

\section*{Competing interests}
We declare we have no competing interests.

\section*{Funding}
T.A.H. is supported by EPSRC (grant EP/Y00857X/1) and also acknowledges travel support from HCMUT\text{-}VNUHCM.

\bibliographystyle{unsrtnat}
\bibliography{mybib}

\clearpage
\begin{center}
  {\LARGE\bfseries Supplementary material}
\end{center}

\appendix
\renewcommand\thesubsection{\thesection.\arabic{subsection}}
\renewcommand\thesubsubsection{\thesubsection.\arabic{subsubsection}}

\counterwithin{figure}{section}
\counterwithin{table}{section}
\counterwithin{equation}{section}

\section{Supplement to the methods}\label{appendix:methodology}

\subsection{Scope of the corpus and the completeness guard}\label{appendix:corpus}

The corpus is a complete factorial crossing of six language models, ten payoff scales, five prompt languages, four ordered persona pairings and ten replicates. The unit of collection is the dyad, one ten-round game between two agents of the same model; it contributes two agent-games, an agent-game being the ten decisions one agent takes in one game. Each of the 300 model-by-language-by-scale cells holds exactly 40 dyads, so the corpus is 12{,}000 dyads, 24{,}000 agent-games and 240{,}000 decisions (table~\ref{tab:S-corpus}). The five models the main text reports carry 250 of those cells, 10{,}000 dyads, 20{,}000 agent-games and 200{,}000 decisions.

Balance is enforced rather than assumed. The analysis build refuses to write any table unless two conditions hold. First, every model-by-language-by-scale cell contains exactly 40 dyads, with none missing and none duplicated. Second, the payoffs recorded in each game recover, to a relative tolerance of $10^{-6}$, the payoff scale that cell was run at, which catches a mislabelled or misfiled run rather than trusting the directory name. The guard exists because the corpus was collected in batches over several sessions and an incomplete batch is otherwise invisible in an aggregate.

The polarity of the action mapping is load-bearing for every cooperation rate in this article, so it was checked twice and in two ways. In the prompt, Option~A pays less in both columns of a penalty matrix that the agent is told to minimise, and is therefore the dominant, defecting action, while mutual Option~B at $2\lam$ beats mutual Option~A at $6\lam$ and is the cooperative one. In the data, the realised penalties recorded for each pair of actions reproduce that matrix. Cooperation means Option~B throughout.

The realised score-action relationships reproduce the intended dilemma structure in the logs: own cooperation correlates with own utility at Spearman $\rho = -0.015$, the opponent's cooperation with own utility at $\rho = +0.916$, and dyad welfare with joint cooperation at $\rho = +0.985$. These are descriptive checks of the recorded payoff mapping and action polarity, not a test of model comprehension.

\subsection{Endpoints, seeds, parsing and fallback}\label{appendix:endpoints}

The six models were served through the Kaggle Benchmarks model proxy on an OpenAI-compatible chat-completions endpoint, at temperature 1.0, with the identifiers listed in table~\ref{tab:A-endpoints}. Four of the six identifiers carry a date and two do not, which matters because an undated identifier can in principle be re-pointed by the provider, and one that is additionally marked as a preview is announced as provisional.

\begin{table}[htbp]
\centering
\footnotesize
\caption{\textbf{Endpoint identifiers.} As addressed through the Kaggle Benchmarks model proxy during collection. The last column records whether the identifier fixes a build. Gemini 3.1 Flash-Lite is reported in \S\ref{appendix:sixth_model} rather than in the main text because its identifier is both undated and marked as a preview.}
\label{tab:A-endpoints}
\begin{tabular}{@{}llc@{}}
\toprule
Model as printed & Endpoint identifier & Dated \\
\midrule
Claude Haiku 4.5      & \texttt{claude-haiku-4-5-20251001}      & yes \\
GPT-5.4 Nano          & \texttt{gpt-5.4-nano-2026-03-17}        & yes \\
Gemini 3.5 Flash-Lite & \texttt{gemini-3.5-flash-lite}          & no  \\
Qwen3 235B-A22B       & \texttt{qwen3-235b-a22b-instruct-2507}  & yes \\
Grok 4.20             & \texttt{grok-4.20-0309-non-reasoning}   & yes \\
Gemini 3.1 Flash-Lite & \texttt{gemini-3.1-flash-lite-preview}  & no, preview \\
\bottomrule
\end{tabular}
\end{table}

The repository records corpus assembly between 2 August and 5 September 2026. Those commit dates bound the archived collection work, but the raw files do not retain per-request timestamps. Exact provider request dates therefore cannot be reconstructed from the release.

The release records matching by game identifier across payoff scales. It does not retain the API seed values, so it supports a matched-block analysis but does not verify provider-specific seed semantics. The robustness analysis therefore permutes payoff-scale labels within recorded game-ID blocks.

With 10,000 blocked permutations, the scale range remains significant for all six models: $p=0.0019$ for Claude Haiku 4.5 and $p=0.0001$ for each of the other five models, the permutation floor. The ranges are unchanged because the analysis only conditions the null on the recorded matching structure.

Decisions are elicited as free text and parsed by the protocol's own matcher. A reply the matcher cannot resolve is regenerated, up to two further attempts, and if all three attempts fail the decision is recorded as Option~A, which under this framing is defection. The direction of the resulting bias is therefore known: the fallback can only push a measured cooperation rate down, never up, and it cannot manufacture the cooperative behaviour reported in the main text. Each collection run checks its own fallback rate against a tolerance of 0.02 once the sweep has finished, and records a failed assertion in its benchmark report if the rate exceeds it. That check flags a run rather than halting it. \rev{For Gemini 3.5 Flash-Lite the per-decision records were retained and give the rates by language and scale (\S D.4); for the other five models} the achieved rates are not carried in the released per-game tables, so we cannot report how they vary with the payoff scale. If the fallback rate were itself to rise with the printed magnitude of the payoffs, it would bias the smallest measured ranges downward rather than create them.

\rev{\paragraph{Cost.} The unit of replication is one model's grid of 2{,}000 dyads and 40{,}000 decisions, which used 17.8 million input and 0.08 million output tokens. At list prices this grid cost US\$5.54 on Gemini 3.5 Flash-Lite, US\$3.68 on GPT-5.4 Nano, US\$5 on Gemini 3.1 Flash-Lite,  US\$34 on Claude Haiku 4.5, US\$67 on Grok 4.20 and US\$13 on Qwen3 235B-A22B. The whole study, including pilot, repeat and failed runs, cost under US\$160.}

\subsection{The scaled payoff matrix at all ten scales}\label{appendix:payoff_scaling}

The manipulation multiplies the four cells of the penalty matrix by a positive constant $\lam$ and changes nothing else in the prompt. With base penalties $T = 0$, $R = 2$, $P = 6$, $S = 10$ the agent is shown the values in table~\ref{tab:A-lambdas}. Two features of the rendering matter for interpretation. The temptation payoff prints as 0 at every scale, so only three of the four numbers move. And the largest number shown is $10\lam$, so every payoff printed is at most 1 exactly when $\lam \leq 0.1$, which two of the ten scales satisfy; that is the sub-unit boundary the persona result in \S3.4 is organised around, and it is a fact about the rendering rather than a free parameter.

\begin{table}[htbp]
\centering
\footnotesize
\caption{\textbf{The payoff matrix as printed at each of the ten scales.} Values are the penalties printed to the agent, which the prompt instructs it to minimise, in the temptation, reward, punishment and sucker notation. The horizontal rule marks the sub-unit boundary, above which some printed payoff exceeds 1.}
\label{tab:A-lambdas}
\begin{tabular}{@{}lcccc@{}}
\toprule
$\lam$ & $T = 0\lam$ & $R = 2\lam$ & $P = 6\lam$ & $S = 10\lam$ \\
\midrule
0.01   & 0 & 0.02 & 0.06 & 0.1 \\
0.1    & 0 & 0.2  & 0.6  & 1 \\
\midrule
0.25   & 0 & 0.5  & 1.5  & 2.5 \\
0.5    & 0 & 1    & 3    & 5 \\
1      & 0 & 2    & 6    & 10 \\
2      & 0 & 4    & 12   & 20 \\
5      & 0 & 10   & 30   & 50 \\
10     & 0 & 20   & 60   & 100 \\
100    & 0 & 200  & 600  & 1000 \\
1000   & 0 & 2000 & 6000 & 10000 \\
\bottomrule
\end{tabular}
\end{table}

\subsection{Prompt templates and a framing artefact}\label{appendix:templates}

The prompt templates are byte-identical to the resource files of the FAIRGAME protocol~\cite{buscemi2025fairgame} and were checked against them before every collection run, by a test that compares the rendered template with the resource file byte for byte. The check exists because the templates carry Arabic, Chinese and Vietnamese text inline, and a single save in a non-UTF-8 encoding silently corrupts the prompt actually sent to the model rather than only a comment.

Artefacts of the published templates are reproduced rather than repaired, so that the corpus matches the protocol as published. The consequential one is that the Arabic and Chinese versions state the four cells as penalties while phrasing the goal sentence as maximising a reward. We therefore treat the five languages as five distinct stimuli rather than as five renderings of one, and we do not attribute a difference in the \emph{level} of cooperation between two languages to the language alone. The wording is fixed within each language across the ten payoff scales, so it does not invalidate the within-language rescaling comparison. It may nevertheless interact with numerical magnitude; consequently, cross-language differences in scale sensitivity cannot be attributed to language alone. Every scale-sensitivity quantity reported in the main text is computed within a language.

The persona line is a single sentence in the language of the prompt, naming the agent's own disposition and never the opponent's. In English the two forms are \emph{You are cooperative.} and \emph{You are selfish.} Because the opponent's persona is never disclosed, a persona pairing reaches a player only through realised behaviour, and not before round 2.

\subsection{The rule base}\label{appendix:rule_based}

Each agent-game is encoded as a sequence of ten two-letter tokens of the form $\langle$previous outcome$\rangle\langle$current action$\rangle$, over the outcome letters $E$ (no history yet), $R$ (both cooperated), $S$ (the player cooperated and was exploited), $T$ (the player exploited a cooperator) and $P$ (both defected), and the actions $C$ and $D$. The vocabulary is therefore eleven symbols including padding, and it is shared exactly between the synthetic training corpus and the language-model transcripts, which is what makes a classifier trained on the former applicable to the latter.

All four canonical strategies are memory-one, so each prescribes an action as a function of the previous outcome letter alone (table~\ref{tab:A-rules}). Deciding whether a trajectory is \emph{exactly} what a strategy would have played on the history that actually occurred is then a per-token lookup, with no learning and no tolerance parameter.

\begin{table}[htbp]
\centering
\footnotesize
\caption{\textbf{The four canonical memory-one rules.} The action each rule prescribes given the previous round's joint outcome. Tit-for-Tat and Win-Stay-Lose-Shift differ only in the two columns $T$ and $P$, which is why ten rounds of mutual cooperation are consistent with AllC, Tit-for-Tat and Win-Stay-Lose-Shift at once.}
\label{tab:A-rules}
\begin{tabular}{@{}lccccc@{}}
\toprule
& $E$ (first move) & $R$ & $S$ & $T$ & $P$ \\
\midrule
AllC & $C$ & $C$ & $C$ & $C$ & $C$ \\
AllD & $D$ & $D$ & $D$ & $D$ & $D$ \\
Tit-for-Tat & $C$ & $C$ & $D$ & $C$ & $D$ \\
Win-Stay-Lose-Shift & $C$ & $C$ & $D$ & $D$ & $C$ \\
\bottomrule
\end{tabular}
\end{table}

Formally, writing $\mathbf{a} = (a_1, \ldots, a_N)$ for the agent's actions and $\mathbf{o} = (o_1, \ldots, o_N)$ for the opponent's, and $\pi_k(\cdot)$ for the action prescribed by rule $k$ in table~\ref{tab:A-rules}, the match is
\[
  \text{consistent}(\mathbf{a}, \mathbf{o}, k)
  \;\equiv\;
  \bigwedge_{t=1}^{N} \bigl[\, a_t = \pi_k(e_t) \,\bigr],
  \qquad
  e_1 = E, \quad
  e_t = \mathrm{outcome}(a_{t-1}, o_{t-1}) \ \ (t \geq 2),
\]
and the deviation count that the main text calls the rule distance is the same expression with the conjunction replaced by a sum of indicators, $d_k = \sum_{t=1}^{N} \mathbb{I}[a_t \neq \pi_k(e_t)]$, reported as $\min_k d_k$, the number of rounds violating the nearest canonical rule.

Because several rules routinely prescribe the same actions on the history that occurred, the matcher returns a \emph{set} rather than a label. A player who cooperated in every round against a cooperator is exactly consistent with AllC, Tit-for-Tat and Win-Stay-Lose-Shift simultaneously, and no method, learned or otherwise, can separate them from that evidence; reporting the set is the correct answer rather than a failure of the method. The three provenances are recorded separately and never merged:

\begin{itemize}
\item \emph{deduced}, where exactly one rule fits, so the label is a deduction;
\item \emph{ambiguous}, where several fit, so the rules fix a candidate set and the classifier only ranks inside it;
\item \emph{unmatched}, where no rule fits, so the label is an attribution the classifier makes alone.
\end{itemize}

There is no confidence threshold anywhere in this read-out, and no composite label is emitted: the multi-label answer is the set itself, and where a single label is needed for a table it is the classifier's ranking restricted to that set. The classifier can therefore never overturn a deduction.

\subsection{Classifier training and validation}\label{appendix:classifier}

The learned branch is trained on a synthetic corpus that contains no language-model output at all. Trajectories of the four canonical strategies are generated against random opponents at four execution-noise levels, balanced at 40{,}320 sequences per strategy per level, giving 161{,}280 sequences at each level. Training uses the zero-noise and 5\% levels pooled, 322{,}560 sequences split 80/20 at random, so the held-out split has 64{,}512 sequences; the 10\% and 20\% levels are never trained on and serve as out-of-distribution tests.

\begin{figure}[htbp]
\centering
\includegraphics[width=\linewidth]{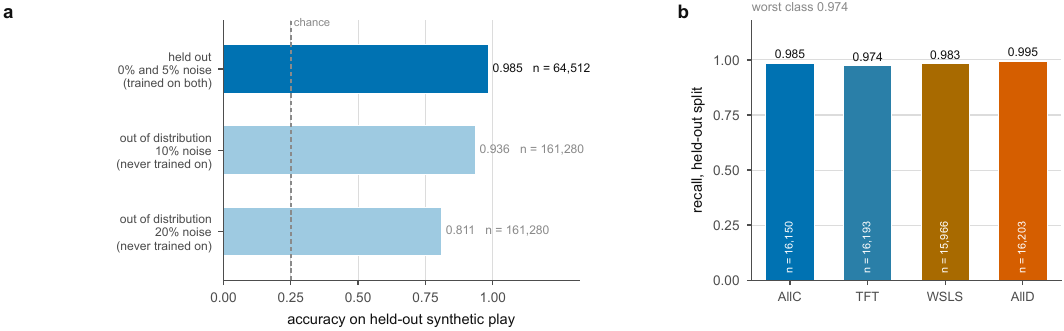}
\caption{\textbf{How well the learned branch recovers a rule that is known by construction.} (a) Accuracy on the held-out split of the training noise levels and on the two noise levels never trained on, against the chance level of a balanced four-way problem. (b) Per-class recall on the held-out split. The two conditional rules are the harder pair: an execution error in either can be read as the signature of the other.}
\label{fig:classifier}
\end{figure}

The network is a single-layer unidirectional LSTM~\cite{lstm}: an embedding of the eleven-symbol vocabulary into 24 dimensions, a recurrent layer of 96 hidden units, dropout 0.15 and a linear map to the four classes. It is optimised with AdamW at learning rate $3\times10^{-3}$ and weight decay $10^{-4}$ under a cosine schedule, for 12 epochs at batch size 512, with seed 1234. Accuracy is 0.985 on the held-out split, 0.936 at 10\% execution noise and 0.811 at 20\%, with per-class recall on the held-out split of 0.985 for AllC, 0.995 for AllD, 0.974 for Tit-for-Tat and 0.983 for Win-Stay-Lose-Shift (figure~\ref{fig:classifier} and table~\ref{tab:S-classifier}). The two conditional rules are the harder pair, as expected: an execution error in either can be read as the signature of the other.

The 20\% level bounds where the read-out is warranted, and \S\ref{appendix:robust_readout} states where the corpus sits relative to it.

\subsection{Resampling and testing}\label{appendix:stats}

Three dependencies rule out treating agent-games as independent: the two agents of a dyad play each other, the ten rounds inside an agent-game are strongly autocorrelated, and the ten replicates of a cell share a payoff scale. Primary inferential intervals therefore use a nonparametric bootstrap that resamples \emph{whole dyads} with replacement, 1{,}500 draws, percentile method~\cite{efron1993introduction}, with a fixed random stream so that draws are paired across cells. Descriptive figure intervals use whole-dyad resampling. Primary per-model and trend regressions cluster standard errors on the dyad. The supplementary scale-by-language robustness model instead clusters on the documented language-by-game-ID blocks to preserve the matched ten-scale ladders. As a robustness check, we also condition the randomisation on the documented game-ID blocks, preserving each complete ten-scale ladder. This sensitivity does not establish common-random-number semantics, which are not retained in the released tables.

The range of a model's cooperation over the ten scales is a maximum minus a minimum and so is positive by construction. Its bootstrap interval quantifies the precision of the range, not its distance from zero, and the test is supplied separately by a permutation. Because the null is a theorem rather than a nuisance hypothesis, the permutation destroys exactly the association under test and leaves the rest of the design as observed: the payoff-scale label is shuffled among whole dyads within language-by-pairing strata, the unit to which the scale was assigned, and the range of the ten scale means is recomputed. With 2{,}000 draws the smallest attainable $p$ value is $1/2001 = 0.0005$. Shuffling agent-games independently would be anti-conservative, because the two agent-games of a dyad would then be free to land in different scale groups although they were generated together.

The variance decomposition is a type-II analysis of variance on cell means with terms for model, language and payoff scale and their three two-way interactions, and no three-way term, so the residual is the omitted three-way interaction: 144 degrees of freedom on the 250 cells of the five main-text models, and 180 on all 300 cells.

\section{Supplement to the results}\label{appendix:results}

\subsection{Cooperation by model and payoff scale}\label{appendix:by_scale}

Table~\ref{tab:S-scale} gives the cooperation rate of every model at every payoff scale, with a bootstrap interval under each entry, for all six models. It provides the model-by-scale estimates underlying figure~2c.

\subsection{Cooperation by model, prompt language and payoff scale}\label{appendix:by_language}

Table~\ref{tab:S-language} gives the same quantity broken down by prompt language, 40 dyads per entry, with the across-scale range in the final column. Over all six models that range runs from 0.080, for Gemini 3.1 Flash-Lite in Arabic, to 0.855, for Grok 4.20 in English. Table~\ref{tab:S-language} provides the underlying model-by-language-by-scale estimates used in figure~3.

\subsection{Variance decomposition}\label{appendix:variance}

Table~\ref{tab:S-variance} is the type-II analysis of variance on the 250 cell means of the five main-text models, and table~\ref{tab:S-variance-all} the same decomposition on all 300 cells. Every term significant in the first is significant in the second and in the same order, and the interaction of the payoff scale with the model again carries several times the variance the scale carries as a main effect.

A third fit bears on how the main effect should be read. Refitting the five-model decomposition with Grok 4.20 removed leaves the payoff-scale main effect not significant, at $F_{9,108} = 1.15$, $p = 0.33$ and $0.9\%$ of the variance across the remaining 200 cells, while every one of those four models still departs from the invariance on its own (table~1). The pooled main effect is therefore carried by a single member of the panel: it is a statement about that model surviving the averaging rather than a property of language models in general, which is why the main text reports it as such rather than as the headline.

\subsection{Persona, in full}\label{appendix:persona_full}

Table~\ref{tab:S-persona} gives the persona effect, the cooperation rate under a cooperative instruction minus that under a selfish one, at each payoff scale and for all six models. The two leftmost columns are the scales at which every printed payoff is at most 1.

Two models invert the instruction at every scale rather than only below the sub-unit boundary. Pooled over the grid, Qwen3 235B-A22B cooperates 0.072 under the cooperative persona and 0.798 under the selfish one, and Grok 4.20 inverts less extremely but throughout. The most likely reading is that the story the prompt tells is ambiguous about whom one is cooperating with, one's partner in crime or the interrogating authority, and that these two models resolve that ambiguity opposite to the convention the payoff matrix encodes. The inversion is not the subject of this article and we make no claim about its cause; we report it because an analysis that quietly excluded the two models whose persona response runs backwards would flatter the persona result. Its magnitude, like everything else measured here, depends on the payoff scale, and the per-scale sequence is not monotone, so no trend is asserted.

\subsection{The strategy read-out, in full}\label{appendix:readout_full}

Table~\ref{tab:S-deduced} gives the share of agent-games matched exactly by one canonical rule, by model and payoff scale, computed by exact matching with no learned component. Table~\ref{tab:S-mix} gives the four-way label composition over the same grid.

Table~\ref{tab:S-provenance} is the one that bounds how the composition may be read. For each label it gives the share of the agent-games carrying it that were deduced, ambiguous or unmatched. The asymmetry is severe: AllC and AllD carry a substantial deduced share, at $23.6\%$ and $35.6\%$ over the corpus, whereas $94.6\%$ of the Tit-for-Tat labels and $97.1\%$ of the Win-Stay-Lose-Shift labels are attributions made where no canonical rule fits at all. A movement in the Tit-for-Tat share is therefore a statement about which canonical rule the play most resembles, and not evidence that any model is running Tit-for-Tat. The two claims the main text makes about rule agreement, the falling deduced share and the rising rule distance, do not depend on the classifier at all.

\subsection{Why the model curves cancel under pooling}\label{appendix:pooling}

Tables~\ref{tab:S-shapes} and~\ref{tab:S-shapes-all} give the pairwise correlation between the ten-point cooperation-against-scale curves of each pair of models, for the five main-text models and for all six.

The cancellation is arithmetic. Writing $c_m(\lam)$ for model $m$'s curve and $\bar{c}(\lam)$ for their average, the range of $\bar{c}$ is bounded above by the average of the individual ranges, with equality only if the curves attain their extremes at the same scales and move together in between. Here six of the ten pairs are anticorrelated and the mean correlation is $-0.10$, so the bound is far from tight: the pooled range is 0.154 against a mean within-model range of 0.249, an attenuation of 1.62. With the sixth model included the same quantities are 0.125, 0.219 and 1.76, with nine of the fifteen pairs negative, so the pooling argument is stronger over the whole corpus than over the five models the main text reports.

This is not a power problem. Collecting more dyads tightens the interval around $\bar{c}$; it does not make $\bar{c}$ a better description of any $c_m$.

\begin{figure}[htbp]
\centering
\includegraphics[width=\linewidth]{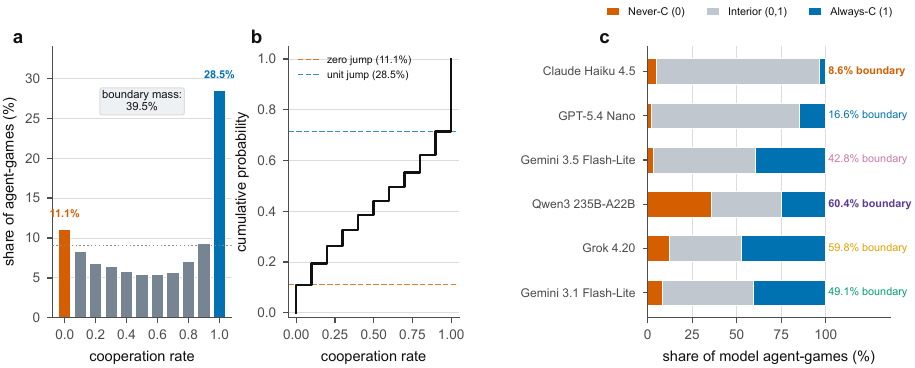}
\caption{\textbf{Bimodal cooperation distribution and model-level policy heterogeneity.} (a) Pooled cooperation distribution across 24{,}000 agent-games, showing extreme boundary inflation: $11.1\%$ of games never cooperate and $28.5\%$ always cooperate, totaling $39.5\%$ boundary mass compared to $18.2\%$ under a flat reference benchmark. (b) Empirical cumulative distribution function (ECDF), displaying the steep vertical steps at the defection and cooperation boundaries. (c) Decomposition into never-cooperate ($c=0$), interior mixed play ($0 < c < 1$), and always-cooperate ($c=1$) across all six models. The boundary share varies seven-fold across models, from $8.6\%$ in Claude Haiku 4.5 (unimodal interior distribution) to $60.4\%$ in Qwen3 235B-A22B and $59.8\%$ in Grok 4.20 (predominantly degenerate pure-policy play).}
\label{fig:coop_dist}
\end{figure}

The cancellation of model curves under pooling reflects an underlying structural divide in how models play: per-game cooperation is not a graded Gaussian quantity, but a mixture over distinct policy regimes (figure~\ref{fig:coop_dist}). Across all 24{,}000 agent-games, $11.1\%$ never cooperate and $28.5\%$ always cooperate, so that two discrete boundary atoms carry $39.5\%$ of all play against $18.2\%$ expected under a flat distribution across the 11 attainable rates.

Crucially, this boundary concentration is a model-level trait rather than an invariant property of the task. In Claude Haiku 4.5, boundary mass accounts for only $8.6\%$ of play; its distribution is unimodal with an interior mode at $0.2$, reflecting an adaptive policy that explores intermediate action frequencies. In stark contrast, Qwen3 235B-A22B ($60.4\%$ boundary mass) and Grok 4.20 ($59.8\%$ boundary mass) play predominantly degenerate pure strategies, switching between always cooperating and always defecting. A single pooled mean cooperation rate therefore conflates two distinct phenomena: how often a model commits to an unconditional pure policy, and which pure policy it selects.

\subsection{The sixth model, Gemini 3.1 Flash-Lite}\label{appendix:sixth_model}

The sixth model was run on the identical grid, with the identical protocol, and every quantity the main text reports for the five is reported for it here and in the tables of this section, which cover all six models throughout.

\paragraph{Why it is reported here rather than in the main text.} Its endpoint
identifier, \texttt{gemini-3.1-\allowbreak flash-lite-\allowbreak preview}, is both undated and marked by the provider as provisional, so what was served under that name while the corpus was collected cannot be re-addressed afterwards and a reader cannot reproduce the run against the same object. We state the cost of drawing the line there rather than leaving it to be found. The identifier of Gemini 3.5 Flash-Lite, which stays in the main text, is undated as well and could in principle be re-pointed to a later build, so it carries a weaker form of the same risk; the line drawn here is between an identifier marked provisional and one that is not, and not between pinned and unpinned. The decision was taken on the identifier and not on the results.

\paragraph{What it does.} Its cooperation range across the ten scales is 0.072
\ci{0.060}{0.133}, the smallest in the corpus, with a permutation $p = 0.0005$ and a clustered Wald statistic of 83.1 on nine degrees of freedom, so it departs from the invariance on the same evidence as the other five. Its across-scale range by language runs from 0.080 in Arabic to 0.170 in Vietnamese. Its persona effect is $+0.577$ below the sub-unit boundary and $+0.680$ above it, a shift of 0.103 \ci{0.075}{0.131} with no sign reversal. Its opening-move range is 3.68 times its range over rounds 2 to 10, the largest ratio in the corpus, so the localisation of the effect at the first move is more pronounced in it than in any main-text model, not less.

\begin{figure}[htbp]
\centering
\includegraphics[width=\linewidth]{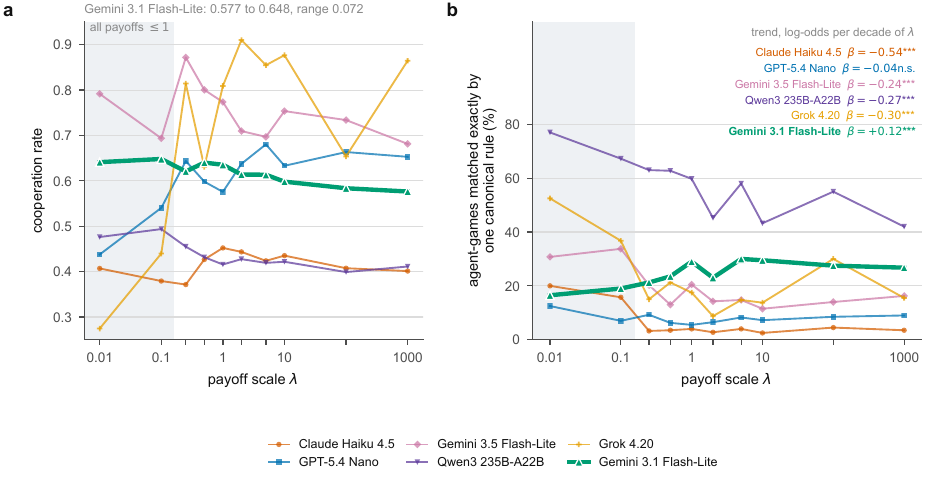}
\caption{\textbf{The sixth model beside the five reported in the main text.} Gemini 3.1 Flash-Lite is drawn heavier and in front. (a) Cooperation against the payoff scale, where it sits inside the spread of the other five. (b) Share of agent-games matched exactly by one canonical rule, with each model's trend per decade of $\lam$. It is the only model of the six whose trend is positive.}
\label{fig:supp_model}
\end{figure}

\paragraph{The one result it reverses.} It is the only model of the six whose
share of agent-games matched exactly by one canonical rule \emph{rises} with the payoff scale, from $17.8\%$ below the sub-unit boundary to $26.3\%$ above it, with a significant positive trend of $\beta = +0.125$ per decade of $\lam$; its rule distance is flat ($\beta = -0.019$, $p = 0.26$); figure~\ref{fig:supp_model} draws it beside the five. Over the whole corpus, the move towards less rule-governed play therefore holds in four models of six, with one showing no trend and one moving the other way.

\subsection{The opening move, in detail}\label{appendix:firstmove_detail}

Table~\ref{tab:S-firstmove} gives, for all six models, the across-scale range of cooperation on the opening move and over rounds 2 to 10, and their ratio. The two windows rest on 400 and 3{,}600 decisions per cell respectively, so their ranges are not variance-matched: a range computed from fewer observations is upward-biased, and the ratios are therefore descriptive rather than tests. The comparison is still informative because the bias runs the same way in every model while the ratio does not, spanning 0.47 in Qwen3 235B-A22B to 3.68 in Gemini 3.1 Flash-Lite.

\begin{figure}[htbp]
\centering
\includegraphics[width=\linewidth]{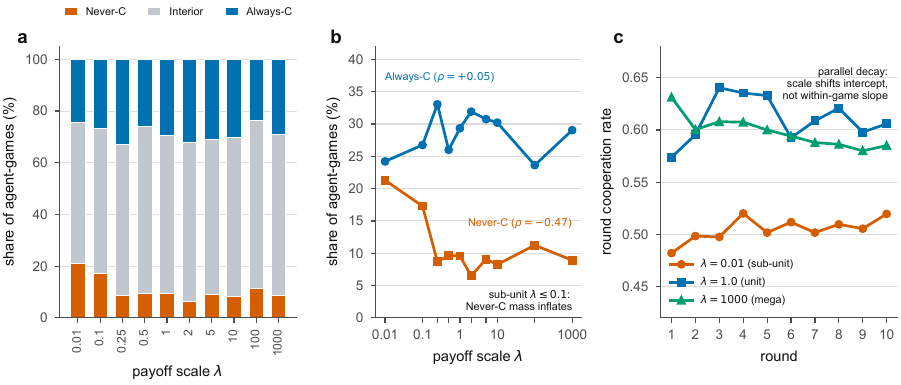}
\caption{\textbf{The scaling mechanism: ex-ante policy re-weighting versus within-game dynamics.} (a) Share of agent-games playing never-cooperate ($c=0$), interior mixed strategies, and always-cooperate ($c=1$) across the ten payoff scales $\lam$. The never-cooperate share drops from $21.3\%$ at $\lam = 0.01$ to $8.9\%$ at $\lam = 1000$. (b) Opposing trajectories of the two boundary atoms against $\log_{10}\lam$: the always-cooperate share rises modestly (Spearman $\rho = +0.05$), while the never-cooperate share falls substantially (Spearman $\rho = -0.47$), showing that the scale effect is concentrated in the relief of unconditional defection at sub-unit stakes. (c) Round-by-round cooperation rate across rounds 1 to 10 for three anchor scales ($\lam = 0.01$, $\lam = 1.0$, and $\lam = 1000$). The similar decay curves are consistent with, but do not establish, an early effect of payoff magnitude.}
\label{fig:mechanism}
\end{figure}

Figure~\ref{fig:mechanism} describes how behaviour changes with the payoff multiplier. Across the ten-scale ladder, the share of agents committing to unconditional defection falls from $21.3\%$ at $\lam = 0.01$ to $8.9\%$ at $\lam = 1000$ (Spearman $\rho = -0.47$), with the sharpest transition occurring across the sub-unit boundary $\lam \leq 0.1$. Meanwhile, the unconditional cooperation share rises modestly from $24.2\%$ to $29.0\%$ (Spearman $\rho = +0.05$).

In contrast, the within-game trajectories across rounds 1 to 10 are similar at the three anchor scales shown in figure~\ref{fig:mechanism}c. Together with the opening-move analysis, this pattern is consistent with an early effect of the printed matrix. It does not distinguish that account from other processes that persist through the game.

\subsection{Outcome fairness, net exploitation and the V-shaped welfare frontier}\label{appendix:fairness_welfare}

Studies evaluating machine behaviour in multi-agent environments often report outcome fairness, measured via the Gini coefficient or payoff disparity between paired agents, interpreting lower disparity as evidence of egalitarian or prosocial alignment. In symmetric normal-form games played between identical agents, however, within-dyad inequality is not an independent behavioural dimension: it is an exact algebraic function of one-directional exploitation.

Because both agents face identical payoff matrices, whenever both choose the same action (mutual defection Option~A or mutual cooperation Option~B), their per-round payoffs are identical by construction ($\pi_1 = \pi_2$). A payoff disparity can only arise in asymmetric rounds where one agent cooperates while the other defects. For this analysis only, utility is expressed on the scale-normalised $[0,1]$ support, $\tilde u=(10\lambda-\mathrm{penalty})/(10\lambda)$, so that an asymmetric round has unit utility disparity. All utility-gap, Gini and welfare quantities below are therefore scale-normalised measures rather than raw penalty differences. Across a ten-round repeated interaction, the relationship is exact:
\begin{equation}
\text{signed\_gap} \;=\; p_{DC} - p_{CD}, \qquad
\text{utility\_gap} \;=\; |p_{DC} - p_{CD}| \;=\; p_{\text{exploit}} - 2 \min(p_{CD}, p_{DC}),
\label{eq:fairness_identity}
\end{equation}
where $p_{DC}$ and $p_{CD}$ denote the proportion of rounds in which the focal agent exploits or is exploited by its partner, respectively. Across all 12{,}000 dyads in the corpus, the maximum residual from this identity is zero. Some $31.9\%$ of dyads have zero utility gap, and the observed gap correlates with total one-directional exploitation at $r = 0.901$. Within-dyad inequality therefore measures coordination failure rather than a distributional preference.

\begin{figure}[htbp]
\centering
\includegraphics[width=\linewidth]{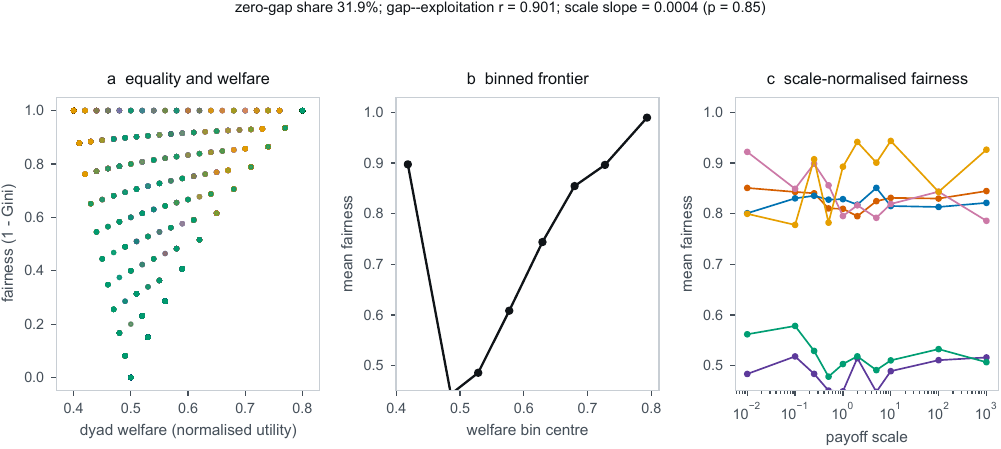}
\caption{\textbf{Outcome fairness, net exploitation and the V-shaped welfare frontier.} (a) Within-dyad fairness ($1 - \text{Gini}$) plotted against joint dyad welfare across all experimental conditions, revealing a characteristic V-shaped relationship where equality peaks at both mutual defection (low welfare) and mutual cooperation (high welfare), with inequality concentrated in the exploitative transition zone. (b) Per-model fairness trajectories showing how different models occupy distinct regimes along the welfare-fairness frontier. (c) Distributional equity across the ten payoff scales: the fitted association is small and imprecise (slope $0.0004$, $p = 0.85$).}
\label{fig:fairness_welfare}
\end{figure}

Figure~\ref{fig:fairness_welfare} illustrates the resulting relationship between fairness and joint welfare. Rather than a monotonic trade-off between equality and efficiency, the empirical frontier is V-shaped: equality is high among low-welfare mutual-defection dyads, falls in the intermediate exploitation region, and returns towards one among high-welfare mutual-cooperation dyads.

Crucially, this structural coordination mechanism explains why the payoff scale exerts zero influence on within-dyad fairness: the regression slopes of both the within-dyad utility gap and Gini coefficient against $\log_{10}\lam$ have slope $0.0004$ ($p = 0.85$). We therefore detect no association between payoff scale and this outcome. This empirical null is distinct from the algebraic identity relating within-dyad inequality to unilateral exploitation.

\begin{table}[htbp]
\centering
\footnotesize
\setlength{\tabcolsep}{4pt}
\caption{\textbf{The corpus, by model.} Every model was run on the same ten payoff scales and five prompt languages. ``Cell $n$'' is the number of dyads in each model $\times$ language $\times$ scale cell; the design is balanced with none missing, and the build refuses to write its tables otherwise. The first five models are the ones the main text reports; Gemini 3.1 Flash-Lite is reported here, for the reason given with the sixth model below.}
\label{tab:S-corpus}
\begin{tabular}{@{}llccrrrc@{}}
\toprule
Model & Scales & Lang. & Dyads & Agent-games & Decisions & Cell $n$ & Coop. \\
\midrule
Claude Haiku 4.5 & 10 & 5 & 2,000 & 4,000 & 40,000 & 40 & 0.415 \\
GPT-5.4 Nano & 10 & 5 & 2,000 & 4,000 & 40,000 & 40 & 0.606 \\
Gemini 3.5 Flash-Lite & 10 & 5 & 2,000 & 4,000 & 40,000 & 40 & 0.751 \\
Qwen3 235B-A22B & 10 & 5 & 2,000 & 4,000 & 40,000 & 40 & 0.435 \\
Grok 4.20 & 10 & 5 & 2,000 & 4,000 & 40,000 & 40 & 0.713 \\
Gemini 3.1 Flash-Lite & 10 & 5 & 2,000 & 4,000 & 40,000 & 40 & 0.617 \\
\midrule
\textbf{All} & 10 & 5 & 12,000 & 24,000 & 240,000 & 40 & 0.590 \\
\bottomrule
\end{tabular}
\end{table}

\begin{table}[htbp]
\centering
\footnotesize
\setlength{\tabcolsep}{4pt}
\caption{\textbf{Cooperation by model and payoff scale.} Each entry aggregates 200 dyads. The bracketed line under each model is a 95\% confidence interval from a bootstrap that resamples whole dyads.}
\label{tab:S-scale}
\resizebox{\textwidth}{!}{%
\begin{tabular}{@{}lcccccccccc@{}}
\toprule
Model & $\lam{=}0.01$ & $\lam{=}0.1$ & $\lam{=}0.25$ & $\lam{=}0.5$ & $\lam{=}1$ & $\lam{=}2$ & $\lam{=}5$ & $\lam{=}10$ & $\lam{=}100$ & $\lam{=}1000$ \\
\midrule
Claude Haiku 4.5 & 0.407 & 0.379 & 0.371 & 0.427 & 0.452 & 0.444 & 0.424 & 0.435 & 0.407 & 0.401 \\
 & \tiny{[0.37,0.45]} & \tiny{[0.34,0.42]} & \tiny{[0.34,0.40]} & \tiny{[0.39,0.46]} & \tiny{[0.42,0.49]} & \tiny{[0.41,0.48]} & \tiny{[0.39,0.46]} & \tiny{[0.40,0.47]} & \tiny{[0.38,0.43]} & \tiny{[0.37,0.43]} \\[2pt]
GPT-5.4 Nano & 0.438 & 0.540 & 0.644 & 0.598 & 0.576 & 0.637 & 0.680 & 0.633 & 0.663 & 0.653 \\
 & \tiny{[0.40,0.48]} & \tiny{[0.50,0.58]} & \tiny{[0.61,0.68]} & \tiny{[0.56,0.63]} & \tiny{[0.54,0.61]} & \tiny{[0.60,0.67]} & \tiny{[0.65,0.71]} & \tiny{[0.60,0.67]} & \tiny{[0.64,0.69]} & \tiny{[0.62,0.68]} \\[2pt]
Gemini 3.5 Flash-Lite & 0.792 & 0.694 & 0.872 & 0.800 & 0.774 & 0.710 & 0.697 & 0.754 & 0.734 & 0.682 \\
 & \tiny{[0.74,0.84]} & \tiny{[0.64,0.74]} & \tiny{[0.85,0.89]} & \tiny{[0.77,0.83]} & \tiny{[0.75,0.80]} & \tiny{[0.67,0.74]} & \tiny{[0.66,0.73]} & \tiny{[0.72,0.78]} & \tiny{[0.70,0.77]} & \tiny{[0.64,0.72]} \\[2pt]
Qwen3 235B-A22B & 0.476 & 0.494 & 0.455 & 0.432 & 0.416 & 0.427 & 0.419 & 0.422 & 0.399 & 0.411 \\
 & \tiny{[0.43,0.52]} & \tiny{[0.44,0.54]} & \tiny{[0.41,0.50]} & \tiny{[0.39,0.47]} & \tiny{[0.38,0.45]} & \tiny{[0.39,0.46]} & \tiny{[0.38,0.45]} & \tiny{[0.39,0.45]} & \tiny{[0.37,0.43]} & \tiny{[0.38,0.44]} \\[2pt]
Grok 4.20 & 0.274 & 0.440 & 0.815 & 0.630 & 0.809 & 0.910 & 0.855 & 0.877 & 0.654 & 0.865 \\
 & \tiny{[0.23,0.32]} & \tiny{[0.39,0.49]} & \tiny{[0.77,0.85]} & \tiny{[0.59,0.67]} & \tiny{[0.77,0.85]} & \tiny{[0.88,0.94]} & \tiny{[0.82,0.89]} & \tiny{[0.84,0.91]} & \tiny{[0.61,0.70]} & \tiny{[0.83,0.90]} \\[2pt]
Gemini 3.1 Flash-Lite & 0.641 & 0.648 & 0.621 & 0.640 & 0.635 & 0.614 & 0.614 & 0.598 & 0.584 & 0.577 \\
 & \tiny{[0.61,0.67]} & \tiny{[0.62,0.68]} & \tiny{[0.59,0.65]} & \tiny{[0.61,0.67]} & \tiny{[0.61,0.67]} & \tiny{[0.58,0.65]} & \tiny{[0.58,0.65]} & \tiny{[0.56,0.63]} & \tiny{[0.54,0.62]} & \tiny{[0.54,0.61]} \\[2pt]
\bottomrule
\end{tabular}
}
\end{table}

\begin{table}[htbp]
\centering
\scriptsize
\setlength{\tabcolsep}{2.5pt}
\caption{\textbf{Cooperation by model, prompt language and payoff scale.} 40 dyads per entry. The final column is the range across the ten scales for that model and language; over all six models it runs from 0.080, for Gemini 3.1 Flash-Lite in Arabic, to 0.855, for Grok 4.20 in English, so the scale sensitivity of a model is a property of the model and the language jointly.}
\label{tab:S-language}
\resizebox{\textwidth}{!}{%
\begin{tabular}{@{}llccccccccccc@{}}
\toprule
Model & Language & 0.01 & 0.1 & 0.25 & 0.5 & 1 & 2 & 5 & 10 & 100 & 1000 & Range \\
\midrule
Claude Haiku 4.5 & English & 0.35 & 0.40 & 0.32 & 0.42 & 0.44 & 0.45 & 0.41 & 0.42 & 0.45 & 0.46 & \textbf{0.136} \\
 & French & 0.65 & 0.47 & 0.50 & 0.33 & 0.37 & 0.43 & 0.41 & 0.40 & 0.38 & 0.40 & \textbf{0.314} \\
 & Vietnamese & 0.17 & 0.17 & 0.28 & 0.29 & 0.26 & 0.20 & 0.30 & 0.22 & 0.23 & 0.23 & \textbf{0.130} \\
 & Chinese & 0.53 & 0.52 & 0.45 & 0.61 & 0.69 & 0.57 & 0.56 & 0.66 & 0.58 & 0.56 & \textbf{0.245} \\
 & Arabic & 0.35 & 0.33 & 0.31 & 0.48 & 0.51 & 0.56 & 0.43 & 0.47 & 0.40 & 0.36 & \textbf{0.249} \\
\addlinespace
GPT-5.4 Nano & English & 0.45 & 0.64 & 0.85 & 0.88 & 0.86 & 0.90 & 0.88 & 0.87 & 0.89 & 0.76 & \textbf{0.449} \\
 & French & 0.75 & 0.78 & 0.78 & 0.57 & 0.55 & 0.60 & 0.62 & 0.67 & 0.55 & 0.74 & \textbf{0.226} \\
 & Vietnamese & 0.49 & 0.46 & 0.58 & 0.47 & 0.50 & 0.62 & 0.75 & 0.54 & 0.65 & 0.63 & \textbf{0.289} \\
 & Chinese & 0.24 & 0.45 & 0.65 & 0.60 & 0.49 & 0.57 & 0.61 & 0.58 & 0.62 & 0.65 & \textbf{0.414} \\
 & Arabic & 0.25 & 0.37 & 0.37 & 0.48 & 0.47 & 0.49 & 0.54 & 0.51 & 0.61 & 0.49 & \textbf{0.360} \\
\addlinespace
Gemini 3.5 Flash-Lite & English & 0.94 & 0.66 & 0.92 & 0.84 & 0.74 & 0.59 & 0.53 & 0.65 & 0.64 & 0.60 & \textbf{0.412} \\
 & French & 0.91 & 0.63 & 0.86 & 0.72 & 0.78 & 0.63 & 0.75 & 0.78 & 0.59 & 0.61 & \textbf{0.317} \\
 & Vietnamese & 0.72 & 0.63 & 0.93 & 0.86 & 0.81 & 0.76 & 0.62 & 0.64 & 0.89 & 0.73 & \textbf{0.314} \\
 & Chinese & 0.69 & 0.80 & 0.76 & 0.64 & 0.70 & 0.73 & 0.67 & 0.76 & 0.66 & 0.73 & \textbf{0.162} \\
 & Arabic & 0.70 & 0.75 & 0.89 & 0.94 & 0.83 & 0.84 & 0.92 & 0.93 & 0.89 & 0.74 & \textbf{0.242} \\
\addlinespace
Qwen3 235B-A22B & English & 0.47 & 0.51 & 0.53 & 0.44 & 0.42 & 0.49 & 0.43 & 0.50 & 0.37 & 0.46 & \textbf{0.155} \\
 & French & 0.41 & 0.40 & 0.35 & 0.39 & 0.37 & 0.36 & 0.35 & 0.35 & 0.34 & 0.33 & \textbf{0.083} \\
 & Vietnamese & 0.52 & 0.55 & 0.48 & 0.48 & 0.45 & 0.45 & 0.47 & 0.44 & 0.40 & 0.42 & \textbf{0.150} \\
 & Chinese & 0.48 & 0.52 & 0.46 & 0.44 & 0.43 & 0.45 & 0.45 & 0.44 & 0.49 & 0.47 & \textbf{0.084} \\
 & Arabic & 0.50 & 0.50 & 0.46 & 0.42 & 0.40 & 0.40 & 0.40 & 0.38 & 0.39 & 0.38 & \textbf{0.124} \\
\addlinespace
Grok 4.20 & English & 0.14 & 0.24 & 0.76 & 0.71 & 0.88 & 0.97 & 0.93 & 1.00 & 0.72 & 1.00 & \textbf{0.855} \\
 & French & 0.46 & 0.53 & 0.99 & 0.74 & 0.80 & 0.92 & 0.89 & 1.00 & 0.62 & 1.00 & \textbf{0.536} \\
 & Vietnamese & 0.26 & 0.74 & 1.00 & 0.72 & 0.99 & 0.99 & 1.00 & 1.00 & 0.72 & 0.98 & \textbf{0.740} \\
 & Chinese & 0.39 & 0.43 & 0.69 & 0.43 & 0.54 & 0.77 & 0.77 & 0.69 & 0.65 & 0.66 & \textbf{0.381} \\
 & Arabic & 0.12 & 0.26 & 0.62 & 0.55 & 0.84 & 0.90 & 0.70 & 0.70 & 0.55 & 0.69 & \textbf{0.786} \\
\addlinespace
Gemini 3.1 Flash-Lite & English & 0.74 & 0.62 & 0.67 & 0.62 & 0.63 & 0.62 & 0.59 & 0.62 & 0.57 & 0.62 & \textbf{0.167} \\
 & French & 0.57 & 0.66 & 0.70 & 0.63 & 0.65 & 0.60 & 0.60 & 0.61 & 0.60 & 0.56 & \textbf{0.145} \\
 & Vietnamese & 0.58 & 0.56 & 0.54 & 0.64 & 0.57 & 0.55 & 0.57 & 0.49 & 0.47 & 0.47 & \textbf{0.170} \\
 & Chinese & 0.62 & 0.66 & 0.53 & 0.61 & 0.59 & 0.60 & 0.60 & 0.58 & 0.58 & 0.58 & \textbf{0.133} \\
 & Arabic & 0.69 & 0.74 & 0.66 & 0.70 & 0.74 & 0.69 & 0.71 & 0.69 & 0.69 & 0.66 & \textbf{0.080} \\
\addlinespace
\bottomrule
\end{tabular}
}
\end{table}

\begin{table}[htbp]
\centering
\footnotesize
\setlength{\tabcolsep}{4pt}
\caption{\textbf{Variance decomposition of the cell means, five models.} Type-II analysis of variance on the 250 model $\times$ language $\times$ scale cell means of the five models the main text reports, where the design is exactly balanced. The payoff scale carries a significant main effect, and its interaction with the model carries three times as much of the variance.}
\label{tab:S-variance}
\begin{tabular}{@{}lrrrrr@{}}
\toprule
Term & Sum sq. & df & $F$ & $p$ & \% variance \\
\midrule
Model & 4.7871 & 4 & 187.27 & $<0.001$ & 45.4 \\
Language & 0.1599 & 4 & 6.25 & $<0.001$ & 1.5 \\
Payoff scale & 0.6119 & 9 & 10.64 & $<0.001$ & 5.8 \\
Model $\times$ language & 1.7901 & 16 & 17.51 & $<0.001$ & 17.0 \\
Model $\times$ scale & 1.8910 & 36 & 8.22 & $<0.001$ & 17.9 \\
Language $\times$ scale & 0.3844 & 36 & 1.67 & $0.018$ & 3.6 \\
Residual & 0.9202 & 144 &  &  & 8.7 \\
\bottomrule
\end{tabular}
\end{table}

\begin{table}[htbp]
\centering
\footnotesize
\setlength{\tabcolsep}{4pt}
\caption{\textbf{Variance decomposition of the cell means, all six models.} The same type-II analysis on all 300 cell means, the sixth model included. Every term significant in the five-model table is significant here and in the same order: the model $\times$ scale interaction again carries several times the variance of the payoff scale as a main effect.}
\label{tab:S-variance-all}
\begin{tabular}{@{}lrrrrr@{}}
\toprule
Term & Sum sq. & df & $F$ & $p$ & \% variance \\
\midrule
Model & 4.8330 & 5 & 170.26 & $<0.001$ & 44.8 \\
Language & 0.1285 & 4 & 5.66 & $<0.001$ & 1.2 \\
Payoff scale & 0.4661 & 9 & 9.12 & $<0.001$ & 4.3 \\
Model $\times$ language & 1.9467 & 20 & 17.15 & $<0.001$ & 18.0 \\
Model $\times$ scale & 2.0646 & 45 & 8.08 & $<0.001$ & 19.1 \\
Language $\times$ scale & 0.3331 & 36 & 1.63 & $0.020$ & 3.1 \\
Residual & 1.0219 & 180 &  &  & 9.5 \\
\bottomrule
\end{tabular}
\end{table}

\begin{table}[htbp]
\centering
\footnotesize
\setlength{\tabcolsep}{4pt}
\caption{\textbf{Persona effect by model and payoff scale.} The cooperation rate of an agent instructed that it is cooperative minus that of an agent instructed that it is selfish, at each scale. The two leftmost columns are the scales at which every payoff printed is at most 1.}
\label{tab:S-persona}
\resizebox{\textwidth}{!}{%
\begin{tabular}{@{}lcccccccccc@{}}
\toprule
Model & $\lam{=}0.01$ & $\lam{=}0.1$ & $\lam{=}0.25$ & $\lam{=}0.5$ & $\lam{=}1$ & $\lam{=}2$ & $\lam{=}5$ & $\lam{=}10$ & $\lam{=}100$ & $\lam{=}1000$ \\
\midrule
Claude Haiku 4.5 & $-0.211$ & $-0.232$ & $+0.206$ & $+0.200$ & $+0.179$ & $+0.261$ & $+0.277$ & $+0.158$ & $+0.168$ & $+0.211$ \\
GPT-5.4 Nano & $-0.172$ & $-0.098$ & $+0.015$ & $-0.026$ & $-0.053$ & $+0.032$ & $+0.041$ & $+0.055$ & $-0.017$ & $+0.100$ \\
Gemini 3.5 Flash-Lite & $-0.335$ & $-0.578$ & $+0.122$ & $+0.234$ & $+0.241$ & $+0.258$ & $+0.313$ & $+0.203$ & $+0.193$ & $+0.230$ \\
Qwen3 235B-A22B & $-0.901$ & $-0.922$ & $-0.811$ & $-0.768$ & $-0.702$ & $-0.635$ & $-0.690$ & $-0.602$ & $-0.657$ & $-0.566$ \\
Grok 4.20 & $-0.454$ & $-0.428$ & $-0.263$ & $-0.484$ & $-0.323$ & $-0.167$ & $-0.235$ & $-0.206$ & $-0.515$ & $-0.234$ \\
Gemini 3.1 Flash-Lite & $+0.552$ & $+0.602$ & $+0.640$ & $+0.670$ & $+0.693$ & $+0.681$ & $+0.707$ & $+0.696$ & $+0.659$ & $+0.692$ \\
\bottomrule
\end{tabular}
}
\end{table}

\begin{table}[htbp]
\centering
\footnotesize
\setlength{\tabcolsep}{4pt}
\caption{\textbf{Share of agent-games matched exactly by one canonical rule (\%).} Computed by exact rule matching, with no learned component. 400 agent-games per entry.}
\label{tab:S-deduced}
\resizebox{\textwidth}{!}{%
\begin{tabular}{@{}lcccccccccc@{}}
\toprule
Model & $\lam{=}0.01$ & $\lam{=}0.1$ & $\lam{=}0.25$ & $\lam{=}0.5$ & $\lam{=}1$ & $\lam{=}2$ & $\lam{=}5$ & $\lam{=}10$ & $\lam{=}100$ & $\lam{=}1000$ \\
\midrule
Claude Haiku 4.5 & 20.0 & 15.8 & 3.2 & 3.5 & 4.0 & 2.8 & 4.0 & 2.5 & 4.5 & 3.5 \\
GPT-5.4 Nano & 12.5 & 7.0 & 9.2 & 6.2 & 5.5 & 6.5 & 8.2 & 7.2 & 8.5 & 9.0 \\
Gemini 3.5 Flash-Lite & 30.8 & 33.8 & 20.2 & 13.0 & 20.5 & 14.2 & 14.8 & 11.5 & 14.0 & 16.2 \\
Qwen3 235B-A22B & 77.0 & 67.2 & 63.0 & 62.7 & 59.8 & 45.2 & 58.0 & 43.2 & 55.0 & 42.0 \\
Grok 4.20 & 52.5 & 36.8 & 15.0 & 21.2 & 17.5 & 8.8 & 14.8 & 13.8 & 30.0 & 15.5 \\
Gemini 3.1 Flash-Lite & 16.5 & 19.0 & 21.2 & 23.5 & 29.0 & 23.0 & 30.0 & 29.5 & 27.5 & 26.8 \\
\bottomrule
\end{tabular}
}
\end{table}

\begin{table}[htbp]
\centering
\scriptsize
\setlength{\tabcolsep}{2.5pt}
\caption{\textbf{Strategy composition by model and payoff scale (\%).} The label assigned by the hybrid read-out: the rule base fixes the candidate set and the LSTM ranks within it, deciding alone only where no canonical rule fits. Columns sum to 100 within each model and scale.}
\label{tab:S-mix}
\resizebox{\textwidth}{!}{%
\begin{tabular}{@{}llcccccccccc@{}}
\toprule
Model & Label & 0.01 & 0.1 & 0.25 & 0.5 & 1 & 2 & 5 & 10 & 100 & 1000 \\
\midrule
Claude Haiku 4.5 & AllC & 27.0 & 20.0 & 14.2 & 20.0 & 21.8 & 18.2 & 15.2 & 18.8 & 12.8 & 14.0 \\
 & TFT & 20.5 & 23.5 & 23.5 & 18.0 & 16.8 & 16.5 & 19.5 & 17.8 & 20.8 & 22.2 \\
 & WSLS & 8.8 & 13.5 & 15.2 & 13.5 & 20.2 & 21.2 & 20.5 & 18.2 & 21.8 & 19.0 \\
 & AllD & 43.8 & 43.0 & 47.0 & 48.5 & 41.2 & 44.0 & 44.8 & 45.2 & 44.8 & 44.8 \\
\addlinespace
GPT-5.4 Nano & AllC & 29.8 & 41.8 & 52.2 & 46.2 & 48.5 & 54.2 & 58.5 & 52.2 & 53.2 & 50.7 \\
 & TFT & 11.8 & 11.5 & 10.5 & 11.8 & 10.2 & 13.2 & 13.8 & 13.2 & 13.0 & 17.8 \\
 & WSLS & 12.5 & 12.5 & 14.5 & 15.5 & 13.8 & 10.5 & 12.5 & 10.2 & 15.5 & 14.0 \\
 & AllD & 46.0 & 34.2 & 22.8 & 26.5 & 27.5 & 22.0 & 15.2 & 24.2 & 18.2 & 17.5 \\
\addlinespace
Gemini 3.5 Flash-Lite & AllC & 73.8 & 59.8 & 74.0 & 64.0 & 58.0 & 60.0 & 51.7 & 59.8 & 56.2 & 51.5 \\
 & TFT & 6.8 & 2.2 & 10.2 & 16.2 & 17.0 & 16.2 & 15.2 & 13.5 & 15.2 & 14.2 \\
 & WSLS & 6.2 & 13.2 & 14.0 & 16.2 & 21.2 & 12.0 & 18.0 & 18.5 & 16.5 & 15.2 \\
 & AllD & 13.2 & 24.8 & 1.8 & 3.5 & 3.8 & 11.8 & 15.0 & 8.2 & 12.0 & 19.0 \\
\addlinespace
Qwen3 235B-A22B & AllC & 42.0 & 47.8 & 39.2 & 38.0 & 36.8 & 36.5 & 35.5 & 35.2 & 34.5 & 33.8 \\
 & TFT & 0.5 & 0.0 & 0.0 & 0.0 & 1.2 & 0.8 & 2.0 & 6.2 & 0.0 & 4.8 \\
 & WSLS & 9.0 & 2.5 & 8.8 & 13.5 & 7.2 & 12.0 & 8.5 & 8.2 & 9.2 & 10.0 \\
 & AllD & 48.5 & 49.8 & 52.0 & 48.5 & 54.8 & 50.7 & 54.0 & 50.2 & 56.2 & 51.5 \\
\addlinespace
Grok 4.20 & AllC & 18.0 & 29.8 & 72.8 & 48.8 & 71.0 & 81.5 & 74.8 & 82.0 & 49.5 & 79.0 \\
 & TFT & 5.8 & 10.0 & 9.5 & 9.0 & 4.2 & 5.8 & 6.8 & 3.2 & 8.2 & 4.2 \\
 & WSLS & 9.8 & 14.8 & 8.5 & 17.5 & 13.5 & 9.0 & 11.5 & 7.0 & 20.0 & 8.0 \\
 & AllD & 66.5 & 45.5 & 9.2 & 24.8 & 11.2 & 3.8 & 7.0 & 7.8 & 22.2 & 8.8 \\
\addlinespace
Gemini 3.1 Flash-Lite & AllC & 55.8 & 57.0 & 58.2 & 61.0 & 59.8 & 55.0 & 58.5 & 53.8 & 52.2 & 52.2 \\
 & TFT & 6.5 & 3.8 & 2.2 & 0.2 & 2.5 & 3.5 & 1.2 & 2.2 & 6.2 & 5.8 \\
 & WSLS & 12.8 & 13.2 & 8.8 & 2.8 & 8.0 & 7.2 & 6.5 & 7.2 & 7.2 & 6.0 \\
 & AllD & 25.0 & 26.0 & 30.8 & 36.0 & 29.8 & 34.2 & 33.8 & 36.8 & 34.2 & 36.0 \\
\addlinespace
\bottomrule
\end{tabular}
}
\end{table}

\begin{table}[htbp]
\centering
\footnotesize
\setlength{\tabcolsep}{4pt}
\caption{\textbf{Provenance of each label.} For every label, the share of the agent-games carrying it that were deduced (exactly one canonical rule fits), ambiguous (several fit, and the LSTM ranked within that set) or unmatched (no canonical rule fits, so the label is an attribution the LSTM makes alone). The asymmetry is severe and bounds how the compositional result may be read: AllC and AllD carry a substantial deduced share, whereas the TFT and WSLS labels are almost entirely attributions.}
\label{tab:S-provenance}
\begin{tabular}{@{}lrccc@{}}
\toprule
Label & $n$ & Deduced (\%) & Ambiguous (\%) & Unmatched (\%) \\
\midrule
AllC & 11,312 & 23.6 & 36.9 & 39.6 \\
TFT & 2,277 & 5.4 & 0.0 & 94.6 \\
WSLS & 2,972 & 2.3 & 0.6 & 97.1 \\
AllD & 7,439 & 35.6 & 0.0 & 64.4 \\
\bottomrule
\end{tabular}
\end{table}

\begin{table}[htbp]
\centering
\footnotesize
\setlength{\tabcolsep}{4pt}
\caption{\textbf{Validation of the LSTM branch.} Trained on synthetic AllC / AllD / Tit-for-Tat / Win-Stay-Lose-Shift trajectories at zero and 5\% execution noise, balanced at 40{,}320 sequences per strategy per level. The 10\% and 20\% noise levels are never trained on. No language-model transcript enters training.}
\label{tab:S-classifier}
\begin{tabular}{@{}lrc@{}}
\toprule
Split & $n$ & Accuracy \\
\midrule
held-out (NoNoise + Noise005) & 64,512 & 0.9846 \\
OOD (Noise01) & 161,280 & 0.9364 \\
OOD (Noise02) & 161,280 & 0.8107 \\
held-out recall: AllC & 16,150 & 0.9853 \\
held-out recall: AllD & 16,203 & 0.9952 \\
held-out recall: TFT & 16,193 & 0.9744 \\
held-out recall: WSLS & 15,966 & 0.9834 \\
\bottomrule
\end{tabular}
\end{table}

\begin{table}[htbp]
\centering
\footnotesize
\setlength{\tabcolsep}{4pt}
\caption{\textbf{Pairwise correlation between the model curves, five models.} Correlation of the ten-point cooperation-against-scale curves of each pair of the five models the main text reports. Six of the ten pairs are negative and the mean is $-0.10$, which is why averaging the curves cancels much of the movement.}
\label{tab:S-shapes}
\resizebox{\textwidth}{!}{%
\begin{tabular}{@{}lccccc@{}}
\toprule
 & Claude Haiku & GPT-5.4 Nano & Gemini 3.5 & Qwen3 235B-A22B & Grok 4.20 \\
\midrule
Claude Haiku 4.5 &  & $+0.11$ & $-0.18$ & $-0.56$ & $+0.40$ \\
GPT-5.4 Nano & $+0.11$ &  & $-0.23$ & $-0.72$ & $+0.85$ \\
Gemini 3.5 Flash-Lite & $-0.18$ & $-0.23$ &  & $+0.22$ & $-0.15$ \\
Qwen3 235B-A22B & $-0.56$ & $-0.72$ & $+0.22$ &  & $-0.71$ \\
Grok 4.20 & $+0.40$ & $+0.85$ & $-0.15$ & $-0.71$ &  \\
\bottomrule
\end{tabular}
}
\end{table}

\begin{table}[htbp]
\centering
\footnotesize
\setlength{\tabcolsep}{4pt}
\caption{\textbf{Pairwise correlation between the model curves, all six models.} The same correlations with the sixth model included: nine of the fifteen pairs are negative and the mean is $-0.08$.}
\label{tab:S-shapes-all}
\resizebox{\textwidth}{!}{%
\begin{tabular}{@{}lcccccc@{}}
\toprule
 & Claude Haiku & GPT-5.4 Nano & Gemini 3.5 & Qwen3 235B-A22B & Grok 4.20 & Gemini 3.1 \\
\midrule
Claude Haiku 4.5 &  & $+0.11$ & $-0.18$ & $-0.56$ & $+0.40$ & $-0.03$ \\
GPT-5.4 Nano & $+0.11$ &  & $-0.23$ & $-0.72$ & $+0.85$ & $-0.70$ \\
Gemini 3.5 Flash-Lite & $-0.18$ & $-0.23$ &  & $+0.22$ & $-0.15$ & $+0.36$ \\
Qwen3 235B-A22B & $-0.56$ & $-0.72$ & $+0.22$ &  & $-0.71$ & $+0.73$ \\
Grok 4.20 & $+0.40$ & $+0.85$ & $-0.15$ & $-0.71$ &  & $-0.57$ \\
Gemini 3.1 Flash-Lite & $-0.03$ & $-0.70$ & $+0.36$ & $+0.73$ & $-0.57$ &  \\
\bottomrule
\end{tabular}
}
\end{table}

\begin{table}[htbp]
\centering
\footnotesize
\setlength{\tabcolsep}{4pt}
\caption{\textbf{Range of cooperation across the ten payoff scales, on the opening move and afterwards.} At round 1 the agent has seen only the payoff matrix and the instructions, so a dependence on the payoff scale there cannot have come through the history of play. The two windows rest on 400 and 3{,}600 decisions per cell, so their ranges are not variance-matched and the ratio is descriptive rather than a test; the bias runs the same way in every model, while the ratio does not.}
\label{tab:S-firstmove}
\begin{tabular}{@{}lccc@{}}
\toprule
Model & Round 1 & Rounds 2--10 & Ratio \\
\midrule
Claude Haiku 4.5 & 0.180 & 0.088 & 2.04 \\
GPT-5.4 Nano & 0.305 & 0.239 & 1.28 \\
Gemini 3.5 Flash-Lite & 0.367 & 0.194 & 1.90 \\
Qwen3 235B-A22B & 0.050 & 0.105 & 0.47 \\
Grok 4.20 & 0.762 & 0.622 & 1.23 \\
Gemini 3.1 Flash-Lite & 0.252 & 0.069 & 3.68 \\
\bottomrule
\end{tabular}
\end{table}

\FloatBarrier

\section{The evolutionary baseline}\label{appendix:egt}

The main text reports what five language models did\rev{, and \S B.7 a sixth}. This appendix computes what the same payoff matrices imply for a population of adaptive agents restricted to the same four canonical rules, on the same ten payoff scales, and compares the two. The purpose is to size a gap rather than to test either object: the baseline says which strategy mixtures are evolutionarily stable when the only sources of variation are payoff magnitude and stochastic execution, and the difference between that and the corpus is \rev{a constructed contrast between two different objects, described rather than tested}. Combining evolutionary modelling with language-model simulation in this way has proved informative in recent work on governance dilemmas~\cite{balabanova2025media,buscemi2025llms}.

\subsection{The model}\label{appendix:egt_model}

We adopt the standard pairwise-comparison process under the small-mutation limit~\cite{sigmund:2010bo,han_when_2021,nowak2004emergence,liu2020evolutionary}, in which a finite well-mixed population of size $Z = 100$ evolves through social imitation governed by the Fermi update rule. The small-mutation limit is the regime in which mutations are rare enough that the population spends almost all of its time in monomorphic states, every individual playing the same strategy, and only occasionally transitions between them; it has a clean analytical solution and is the standard reference baseline in finite-population evolutionary game theory, and it has proved able to reproduce features of human experimental data~\cite{rand2013evolution,zisis2015generosity}. Selection intensity is set to $\beta = 0.1$.  \rev{Because payoffs enter the Fermi rule only through the product $\beta\lam$, the baseline at scale $\lam$ and intensity $\beta$ is the baseline at scale 1 and intensity $\beta\lam$: in this model the payoff scale acts as the intensity of selection, and the five values of $\beta$ in \S\ref{appendix:robust_egt} shift the same curve along the $\lam$ axis.}

To put the analysis on the same incentive structure as the experiments, the FAIRGAME penalty matrix is negated, so that lower-is-better penalties become higher-is-better utilities. The penalties $T_{\text{pen}} = 0$, $R_{\text{pen}} = 2$, $P_{\text{pen}} = 6$, $S_{\text{pen}} = 10$ map to utilities $T = 0$, $R = -2$, $P = -6$, $S = -10$, which satisfy the standard prisoner's dilemma ordering $T > R > P > S$; the transformation is a strict sign flip and preserves the strategic ordering. The whole matrix is then multiplied by $\lam$, exactly as in the experimental protocol, so that the two sides of the comparison are anchored to the same per-round magnitudes.

For a resident strategy $i$ and a mutant $j$, the expected per-round payoff $\pi_{ij}$ is obtained analytically over the same horizon the experiments use, $r = 10$ rounds, by propagating the joint distribution over the two players' actions forward round by round and averaging; no simulation enters. The fixation probability of a single $j$-mutant in a resident $i$-population is
\begin{equation}
\rho_{ji} \;=\; \left(1 + \sum_{k=1}^{Z-1}\prod_{m=1}^{k}
   \frac{T^{-}_{ji}(m)}{T^{+}_{ji}(m)}\right)^{-1},
\label{eq:fixation}
\end{equation}
where $T^{\pm}_{ji}(m)$ are the Fermi-weighted transition rates at composition $m$~\cite{szabo1998evolutionary}. Assembling these probabilities gives an embedded Markov chain over the four monomorphic states whose unique stationary distribution $\sigma$ is the long-run frequency of each strategy in the rare-mutation regime~\cite{sigmund:2010bo}. The reference implementation is the open-source EGTtools library~\cite{domingos2023egttools}.

\subsection{Deterministic continuous dynamics and simplex orbits}\label{appendix:egt_simplex}

In the infinite-population deterministic limit, evolutionary dynamics is described by the standard replicator equation~\cite{sigmund:2010bo,nowak2006five}:
\begin{equation}
\dot{x}_i \;=\; x_i \left[ (\mathbf{A}\mathbf{x})_i - \mathbf{x}^T \mathbf{A} \mathbf{x} \right],
\label{eq:replicator}
\end{equation}
where $\mathbf{x} = (x_1, \dots, x_4)^T$ is the population state on the three-simplex $\Delta_3$ spanned by the four canonical rules (AllC, AllD, TFT, WSLS), and $\mathbf{A}$ is the $4 \times 4$ payoff matrix. Under a scalar payoff transformation $\mathbf{A}' = \lam \mathbf{A}$ with $\lam > 0$, the vector field scales uniformly:
\begin{equation}
\dot{x}_i' \;=\; \lam \, \dot{x}_i.
\label{eq:replicator_scaled}
\end{equation}
Multiplying the payoff matrix by a positive constant multiplies the replicator vector field by that constant. This changes the velocity of state evolution along trajectories by a uniform time reparametrisation $d\tau = \lam \, dt$, but leaves every trajectory, orbit geometry, invariant manifold, and rest point strictly invariant.

\begin{figure}[!htbp]
\centering
\includegraphics[width=\linewidth]{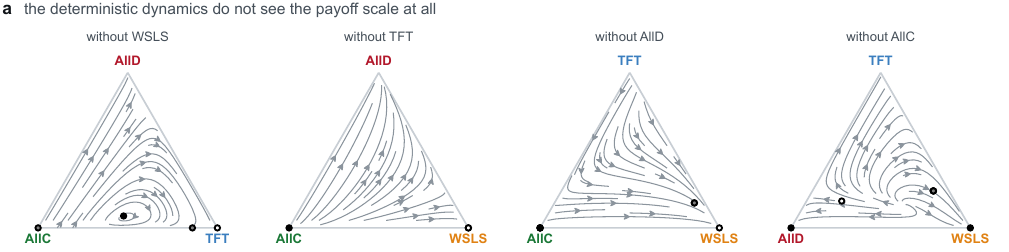}
\caption{\textbf{The deterministic dynamics do not see the payoff scale.} Replicator flow on each of the four faces of the three-simplex spanned by the canonical rules, one face per strategy left out, at $\lam = 1$, $\varepsilon = 0.05$ and ten rounds. Filled circles are stable rest points, open circles unstable. Only one row is drawn because rescaling the payoffs multiplies the field by a constant and moves nothing: the matrix at $\lam = 1000$ is exactly $10^{5}$ times the matrix at $\lam = 0.01$, and over 2{,}000 random population states the normalised fields agree to $4 \times 10^{-14}$.}
\label{fig:app_simplex}
\end{figure}

Figure~\ref{fig:app_simplex} illustrates this mathematical invariance across the four faces of the three-simplex spanned by the canonical rules, leaving one rule out per face. Across 2{,}000 randomly sampled population states, the normalised vector fields at $\lam = 0.01$ and $\lam = 1000$ agree to within $4 \times 10^{-14}$, confirming that continuous deterministic selection cannot account for any empirical dependence on payoff scale.

\subsection{The strategies under execution noise}\label{appendix:egt_strategies}

The four canonical rules enter as memory-one policies with a per-round implementation error $\varepsilon$, the probability of executing the opposite of the intended action. Writing $\mathrm{hi} = 1 - \varepsilon$ and $\mathrm{lo} = \varepsilon$, the probability of cooperating is $\mathrm{hi}$ wherever table~\ref{tab:A-rules} prescribes $C$ and $\mathrm{lo}$ wherever it prescribes $D$, including on the opening move, so AllC, Tit-for-Tat and Win-Stay-Lose-Shift open at $\mathrm{hi}$ and AllD at $\mathrm{lo}$. The rules are therefore the same objects the empirical read-out matches against, which is what makes the two sides of the comparison commensurable.

\subsection{Numerical method and precision audit}\label{appendix:egt_numerics}

The computation is numerically delicate at the top of the payoff grid, and we report the audit rather than only the result. Equation~\eqref{eq:fixation} accumulates a product of $Z-1$ Fermi ratios whose exponents scale with $\beta \lam \Delta\pi$; at the largest scales that product overflows the range of double precision, every fixation probability rounds to exactly 0 or 1, the embedded chain acquires more than one absorbing state, and the eigenvector for eigenvalue 1 is no longer unique. A double-precision solver then returns an arbitrary member of a degenerate eigenspace, which is a wrong answer that carries no warning.

The published grid is therefore computed twice: once in arbitrary precision with \texttt{mpmath} at 200 decimal digits, which is the primary solver, and once in double precision through EGTtools, which serves as a check. Each cell of the grid records the maximum absolute discrepancy between the two solvers and the number of absorbing states the double-precision chain possesses, so a reader can see exactly where double precision fails and by how much.

\subsection{The grid}\label{appendix:egt_grid}

Figure~\ref{fig:egt_grid} draws the stationary frequency of each canonical strategy against the payoff scale at every execution-noise level, and the tables below give the same grid in numbers, together with the precision audit.

\subsection{Finite-population invasion structure across scales}\label{appendix:egt_invasion}

In contrast to the continuous deterministic limit, the payoff scale $\lam$ acts directly on finite populations governed by stochastic imitation. Under the Fermi update rule, the probability that an individual adopting strategy $i$ switches to strategy $j$ upon pairwise comparison is:
\begin{equation}
P(i \to j) \;=\; \frac{1}{1 + \exp\left(-\beta (\pi_j - \pi_i)\right)}
\;=\; \frac{1}{1 + \exp\left(-\beta \lam (\pi_j^{(0)} - \pi_i^{(0)})\right)},
\label{eq:fermi_scaling}
\end{equation}
where $\pi_j^{(0)}$ denotes baseline unscaled payoffs. The payoff scale $\lam$ and the selection intensity $\beta$ enter the update rule exclusively through their product $\beta \lam$. Consequently, scaling the payoffs by $\lam$ is formally equivalent to scaling the selection pressure.

\begin{figure}[!htbp]
\centering
\includegraphics[width=\linewidth]{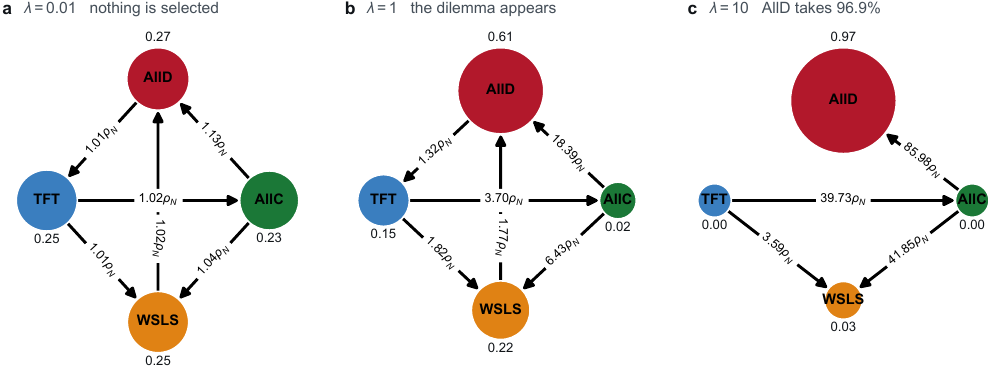}
\caption{\textbf{Where the payoff scale does act: the invasion structure of a finite population.} Nodes are the four monomorphic states, node area is the small-mutation-limit stationary probability, and an edge is drawn where a single mutant fixes more often than neutral drift $1/Z$ would carry it, labelled in multiples of that drift. $Z = 100$, Fermi pairwise comparison, $\beta = 0.1$, $\varepsilon = 0.05$. Because $\lam$ and $\beta$ enter the update rule only through their product, raising the scale is raising selection pressure. The panels stop at $\lam = 10$ because the float64 fixation probabilities underflow beyond it; those cells are solved at 200 digits instead.}
\label{fig:app_invasion}
\end{figure}

Figure~\ref{fig:app_invasion} shows the resulting invasion graphs across the four monomorphic states as $\lam$ varies. At $\lam = 0.01$, selection is negligible, and transition probabilities approach neutral drift ($1/Z = 0.01$). As $\lam$ increases to baseline ($\lam = 1.0$) and high stakes ($\lam = 10.0$), selection sharpens, making AllD an evolutionary sink that resists invasion from all three cooperative rules and fixes with high probability.

\begin{table}[htbp]
\centering
\footnotesize
\caption{\textbf{Stationary distribution of the evolutionary baseline at $\varepsilon = 0.05$.} Long-run frequency of each canonical rule under the pairwise-comparison process in the small-mutation limit, $Z = 100$, $\beta = 0.1$, $r = 10$ rounds, computed in 200-digit arithmetic. The last two columns are the precision audit: the largest absolute disagreement between the 200-digit solver and a double-precision one, and the number of absorbing states the double-precision chain acquires where that number exceeds one. Double precision fails exactly where the audit says it does, at the top of the grid, and returns a wrong answer there without any warning.}
\label{tab:C-egt-eps05}
\begin{tabular}{@{}lcccccc@{}}
\toprule
& \multicolumn{4}{c}{stationary frequency} & \multicolumn{2}{c}{audit} \\
\cmidrule(lr){2-5} \cmidrule(lr){6-7}
$\lam$ & AllC & AllD & TFT & WSLS & $\max|\Delta|$ & absorbing \\
\midrule
0.01 & 0.228 & 0.267 & 0.251 & 0.253 & 4.8e-15 & 0 \\
0.1 & 0.100 & 0.382 & 0.260 & 0.258 & 6.0e-15 & 0 \\
0.25 & 0.049 & 0.446 & 0.258 & 0.247 & 3.6e-15 & 0 \\
0.5 & 0.032 & 0.507 & 0.224 & 0.237 & 1.5e-14 & 0 \\
1 & 0.022 & 0.606 & 0.150 & 0.223 & 1.7e-12 & 0 \\
2 & 0.013 & 0.738 & 0.068 & 0.181 & 1.9e-07 & 0 \\
5 & 0.002 & 0.894 & 0.010 & 0.094 & 4.1e-08 & 0 \\
10 & $2\!\times\!10^{-4}$ & 0.969 & 0.001 & 0.030 & 1.1e-12 & 0 \\
100 & $4\!\times\!10^{-26}$ & 1.000 & $7\!\times\!10^{-26}$ & $4\!\times\!10^{-12}$ & 1.0e+00 & 2 \\
1000 & $2\!\times\!10^{-245}$ & 1.000 & $4\!\times\!10^{-245}$ & $4\!\times\!10^{-112}$ & 1.0e+00 & 2 \\
\bottomrule
\end{tabular}
\end{table}

\begin{table}[htbp]
\centering
\footnotesize
\caption{\textbf{Stationary frequency of Always Defect, across the payoff grid and every execution-noise level.} The same process as table~\ref{tab:C-egt-eps05}. Raising the payoff scale hands the population to Always Defect at every noise level, and does so sooner the noisier execution is. The one exception to monotonicity is the error-free corner $\varepsilon = 0$, where the share dips slightly before rising: with no execution error Tit-for-Tat loses only its opening move to Always Defect, so raising the scale first sharpens selection in Tit-for-Tat's favour. Any positive error rate removes the exception, because it lets Always Defect exploit the retaliation lag repeatedly rather than once.}
\label{tab:C-egt-alld}
\begin{tabular}{@{}lccccc@{}}
\toprule
& \multicolumn{5}{c}{execution error $\varepsilon$} \\
\cmidrule(lr){2-6}
$\lam$ & 0 & 0.05 & 0.1 & 0.2 & 0.3 \\
\midrule
0.01 & 0.263 & 0.267 & 0.269 & 0.268 & 0.263 \\
0.1 & 0.319 & 0.382 & 0.416 & 0.429 & 0.394 \\
0.25 & 0.315 & 0.446 & 0.546 & 0.634 & 0.602 \\
0.5 & 0.304 & 0.507 & 0.661 & 0.811 & 0.822 \\
1 & 0.283 & 0.606 & 0.802 & 0.939 & 0.965 \\
2 & 0.273 & 0.738 & 0.937 & 0.993 & 0.999 \\
5 & 0.347 & 0.894 & 0.998 & 1.000 & 1.000 \\
10 & 0.563 & 0.969 & 1.000 & 1.000 & 1.000 \\
100 & 1.000 & 1.000 & 1.000 & 1.000 & 1.000 \\
1000 & 1.000 & 1.000 & 1.000 & 1.000 & 1.000 \\
\bottomrule
\end{tabular}
\end{table}

\begin{table}[htbp]
\centering
\footnotesize
\setlength{\tabcolsep}{4pt}
\caption{\textbf{The evolutionary baseline beside the corpus, at matched payoff scales.} Left, the stationary frequency of each canonical rule at $\varepsilon = 0.05$; right, the share of agent-games carrying each label, pooled over the five models the main text reports. The two move in opposite directions: the baseline's Always Defect share rises from 0.267 to 1.000 across the grid while the corpus's falls from 0.436 to 0.283.}
\label{tab:C-egt-vs-llm}
\begin{tabular}{@{}lcccccccc@{}}
\toprule
& \multicolumn{4}{c}{evolutionary baseline} & \multicolumn{4}{c}{corpus} \\
\cmidrule(lr){2-5} \cmidrule(lr){6-9}
$\lam$ & AllC & AllD & TFT & WSLS & AllC & AllD & TFT & WSLS \\
\midrule
0.01 & 0.228 & 0.267 & 0.251 & 0.253 & 0.381 & 0.436 & 0.091 & 0.092 \\
0.1 & 0.100 & 0.382 & 0.260 & 0.258 & 0.398 & 0.395 & 0.094 & 0.113 \\
0.25 & 0.049 & 0.446 & 0.258 & 0.247 & 0.505 & 0.266 & 0.107 & 0.122 \\
0.5 & 0.032 & 0.507 & 0.224 & 0.237 & 0.434 & 0.303 & 0.110 & 0.152 \\
1 & 0.022 & 0.606 & 0.150 & 0.223 & 0.472 & 0.277 & 0.099 & 0.152 \\
2 & 0.013 & 0.738 & 0.068 & 0.181 & 0.501 & 0.265 & 0.105 & 0.130 \\
5 & 0.002 & 0.894 & 0.010 & 0.094 & 0.471 & 0.272 & 0.114 & 0.142 \\
10 & $2\!\times\!10^{-4}$ & 0.969 & 0.001 & 0.030 & 0.496 & 0.271 & 0.108 & 0.124 \\
100 & $4\!\times\!10^{-26}$ & 1.000 & $7\!\times\!10^{-26}$ & $4\!\times\!10^{-12}$ & 0.412 & 0.307 & 0.114 & 0.166 \\
1000 & $2\!\times\!10^{-245}$ & 1.000 & $4\!\times\!10^{-245}$ & $4\!\times\!10^{-112}$ & 0.458 & 0.283 & 0.127 & 0.133 \\
\bottomrule
\end{tabular}
\end{table}

\subsection{Comparison with the corpus}\label{appendix:egt_vs_llm}

\begin{figure}[htbp]
\centering
\includegraphics[width=\linewidth]{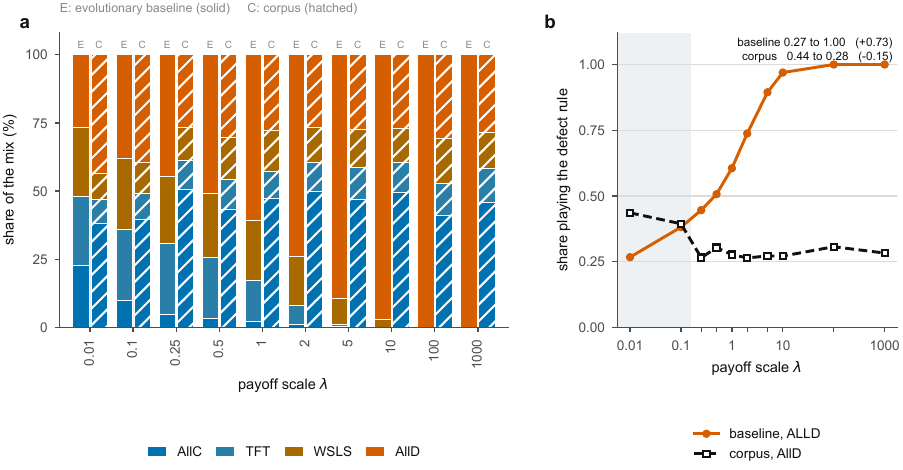}
\caption{\textbf{The evolutionary baseline against the corpus, at matched payoff scales.} (a) Stationary frequency of each canonical strategy under the small-mutation limit ($Z = 100$, $\beta = 0.1$, $r = 10$, $\varepsilon = 0.05$), beside the pooled label composition of the five main-text models at the same payoff scale. (b) The share playing Always Defect alone, the one series on which the two disagree in sign: across the ten scales the baseline rises from 0.27 to 1.00 while the corpus falls from 0.44 to 0.28.}
\label{fig:egt_vs_llm}
\end{figure}

\begin{figure}[htbp]
\centering
\includegraphics[width=\linewidth]{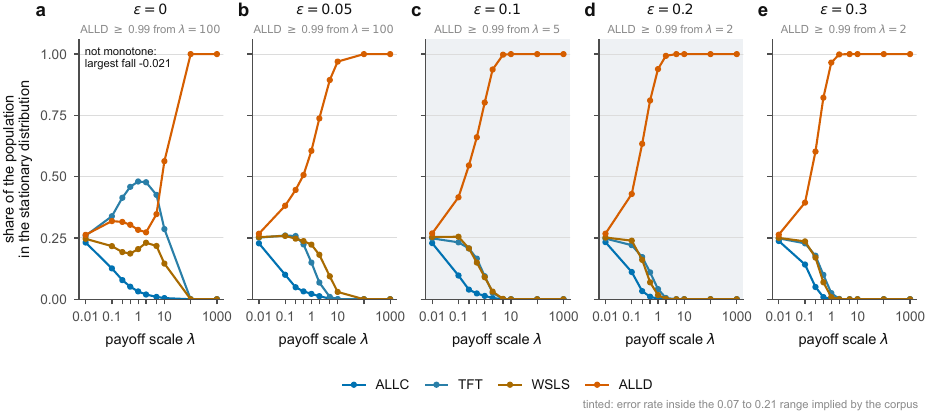}
\caption{\textbf{The evolutionary baseline across the payoff grid and the execution-noise levels.} Stationary frequency of each of the four canonical strategies against the payoff scale, one panel per execution-error rate $\varepsilon$. Always Defect rises with $\lam$ at every noise level, and does so sooner the noisier execution is; the rise is monotone at every positive $\varepsilon$ and dips slightly before rising in the error-free corner $\varepsilon = 0$ (table~\ref{tab:C-egt-alld}).}
\label{fig:egt_grid}
\end{figure}

Figure~\ref{fig:egt_vs_llm} places the stationary distribution beside the pooled label composition of the corpus at each payoff scale.

Two features agree. Both put their greatest behavioural diversity at the smallest payoff scales, where the baseline is close to neutral drift because $\beta \lam \Delta\pi$ is small and the corpus shows its models at their most dissimilar. And in both, a change in $\lam$ moves mass between the conditional cooperators and unconditional defection rather than reshuffling the four strategies arbitrarily.

The direction is the disagreement, \rev{and the comparison is built to show it: a constructed contrast between two different objects, not a test of one against the other.} The baseline drives Always Defect upward as the payoff scale grows, because a larger $\lam$ multiplies the payoff differences that the Fermi rule responds to and so sharpens selection against any rule that can be exploited; at $\varepsilon = 0.05$ its share runs from 0.267 at $\lam = 0.01$ to 1.000 by $\lam = 100$, and the process ends with the population entirely on Always Defect (table~\ref{tab:C-egt-vs-llm}). The corpus does the opposite in aggregate: its pooled AllD share falls from $43.6\%$ at $\lam = 0.01$ to $28.3\%$ at $\lam = 10^{3}$, and its AllC share rises from $38.1\%$ to $45.8\%$. \rev{Given the same matrices and restricted to the same four rules, the two objects move in opposite directions.}

One qualification belongs with the baseline rather than with the comparison. The rise of Always Defect in $\lam$ is monotone at every positive execution error, but not in the error-free corner $\varepsilon = 0$, where the share falls from 0.319 at $\lam = 0.1$ to 0.273 at $\lam = 2$, a dip of 0.046, before rising (table~\ref{tab:C-egt-alld}). The reason is specific and worth stating, because it is a property of the model and not of the arithmetic. With no execution error, AllC, Tit-for-Tat and Win-Stay-Lose-Shift earn exactly $-2\lam$ against each other and against themselves, so the three cooperative rules are neutral among themselves and the ordering is decided entirely by their exchanges with Always Defect. Over ten rounds Tit-for-Tat loses only its opening move to Always Defect, whereas Win-Stay-Lose-Shift keeps being re-exploited and AllC is exploited throughout, so raising $\lam$ first sharpens selection in Tit-for-Tat's favour before Always Defect's one-round advantage finally decides it. Any positive error rate removes the dip, because it lets Always Defect exploit the retaliation lag repeatedly rather than once. The empirically implied error rates are all positive and well above zero (\S\ref{appendix:egt_calibration}), so the qualification does not touch the comparison.

A second disagreement is structural rather than directional. The baseline is language-blind by construction: there is nothing in equation~\eqref{eq:fixation} for a prompt language to enter. In the corpus, the interaction of the model with the prompt language carries $17.0\%$ of the variance across cells, more than the payoff scale carries as a main effect. Whatever the corpus is responding to, a payoff matrix is not a complete description of it.

\rev{We read the gap as a descriptive contrast and not as a falsification. The baseline describes selection among rules and the corpus describes fixed language-model agents, and the prompt effects discussed in the main text could produce the gap as well as any preference the agents hold. The baseline is a reference point for studies of language models in strategic settings, not a quantity this design can explain.}

\subsection{Execution noise is not a free parameter here}\label{appendix:egt_calibration}

In a study with no empirical read-out, $\varepsilon$ would have to be chosen. Here it can be estimated, because the rule base already measures how far each model's play sits from the nearest canonical rule. Dividing a model's mean rule distance by the ten rounds of a game gives a per-round rate at which its realised play departs from the rule it most resembles, which is the empirical analogue of an implementation error (table~\ref{tab:A-egtcalib}).

\begin{table}[htbp]
\centering
\footnotesize
\caption{\textbf{Execution noise implied by the read-out.} The mean number of rounds, out of ten, by which an agent-game departs from the nearest canonical rule, and that quantity as a per-round rate. The two models whose play is least rule-governed imply a noise level at the very edge of the range the classifier was validated on, which is why \S\ref{appendix:robust_readout} treats their strategy labels as the weakest evidence in this article.}
\label{tab:A-egtcalib}
\begin{tabular}{@{}lcc@{}}
\toprule
Model & Mean rule distance (rounds of 10) & Implied $\varepsilon$ \\
\midrule
Claude Haiku 4.5      & 2.073 & 0.207 \\
GPT-5.4 Nano          & 2.120 & 0.212 \\
Gemini 3.5 Flash-Lite & 1.077 & 0.108 \\
Qwen3 235B-A22B       & 0.737 & 0.074 \\
Grok 4.20             & 0.725 & 0.073 \\
Gemini 3.1 Flash-Lite & 1.030 & 0.103 \\
\bottomrule
\end{tabular}
\end{table}

The implied rates span 0.073 to 0.212, so no single noise level matches the whole corpus, and the honest matched comparison for two of the five main-text models is at $\varepsilon = 0.2$ rather than at the 0.05 a study of this kind would ordinarily adopt. The grid in \S\ref{appendix:egt_grid} therefore carries that level, and the conclusion of \S\ref{appendix:egt_vs_llm} is unchanged across it: the direction of the baseline in $\lam$ does not depend on $\varepsilon$.

This mapping should not be over-read. A symmetric per-action flip probability is a clean theoretical object and a coarse abstraction of what makes a language model's play depart from a rule, which includes prompt sensitivity, end-game reasoning and the heterogeneity documented throughout this article. The calibration bounds the comparison; it does not fit the model.

\section{Robustness}\label{appendix:robustness}

\subsection{What including the sixth model changes}\label{appendix:robust_sixth}

Every pooled quantity in this article is reported twice, once on the five main-text models and once on all six, so that the effect of the split can be seen rather than assumed. Including Gemini 3.1 Flash-Lite moves the pooled range from 0.154 to 0.125, the mean within-model range from 0.249 to 0.219 and the attenuation from 1.62 to 1.76; the share of anticorrelated pairs rises from six of ten to nine of fifteen; and in the variance decomposition the payoff scale falls from $5.8\%$ to $4.3\%$ as a main effect while its interaction with the model rises from $17.9\%$ to $19.1\%$ (tables~\ref{tab:S-variance},~\ref{tab:S-variance-all},~\ref{tab:S-shapes} and~\ref{tab:S-shapes-all}).

Every sign, every significance and every ordering agrees between the two, and the pooling argument the main text makes is stronger over all six models than over the five it reports. The single count that the sixth model changes is the strategy-trend tally, and it changes it in the direction that weakens the paper (appendix~\ref{appendix:sixth_model}).

\subsection{Robustness of the evolutionary baseline across 80 settings}\label{appendix:robust_egt}

The finding that the evolutionary baseline predicts an increase in defection with payoff scale is not an artefact of the specific baseline parameters ($Z = 100$, $\beta = 0.1$, $\varepsilon = 0.05$). To confirm this, we solved the stationary distribution across a comprehensive grid of 80 parameter combinations: four population sizes $Z \in \{50, 100, 200, 500\}$, five selection intensities $\beta \in \{0.01, 0.05, 0.1, 0.5, 1.0\}$, and four execution noise levels $\varepsilon \in \{0, 0.05, 0.1, 0.2\}$, across all ten payoff scales (800 cells in total, each computed at 200 decimal digits).

\begin{figure}[!htbp]
\centering
\includegraphics[width=\linewidth]{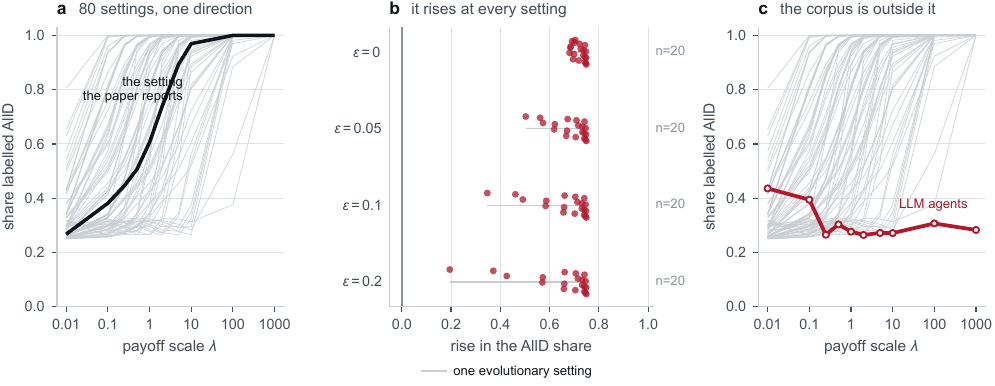}
\caption{\textbf{The baseline's direction is a property of the model, not of its parameters.} (a) The AllD share against the payoff scale, one hairline per evolutionary setting, with the setting reported in the text drawn heavy. (b) The rise in the AllD share from the smallest scale to the largest, by execution error. (c) The corpus on the same axis as the family. Eighty settings, $Z \in \{50, 100, 200, 500\}$ crossed with $\beta \in \{0.01, 0.05, 0.1, 0.5, 1\}$ and $\varepsilon \in \{0, 0.05, 0.1, 0.2\}$, ten payoff scales, 800 cells each solved at 200 significant digits. The rise is monotone at every positive $\varepsilon$; all 19 exceptions are at $\varepsilon = 0$.}
\label{fig:app_robustness}
\end{figure}

Figure~\ref{fig:app_robustness} presents this parameter sweep. Across all 80 settings, the stationary share of Always Defect increases with $\lam$, exceeding $90\%$ at large scales in every setting, with a median increase of 0.716 and a minimum increase of 0.195. In 61 of the 80 settings the rise is strictly monotone; the 19 non-monotone cases occur exclusively at $\varepsilon = 0$, where a transient dip occurs before selection drives AllD to fixation (\S\ref{appendix:egt_vs_llm}). The pooled empirical corpus moves in the opposite direction to this theoretical prediction. This pooled reversal is driven primarily by GPT and Grok; per-model label trajectories are heterogeneous.

\subsection{The read-out at the edge of its validated range}\label{appendix:robust_readout}

The classifier was validated to $20\%$ execution noise, which is two rounds in ten. Table~\ref{tab:A-egtcalib} shows that Claude Haiku 4.5 and GPT-5.4 Nano sit at 2.07 and 2.12 rounds from the nearest canonical rule on average, so for those two models the read-out is operating at the very edge of the range it was validated on, and their label compositions are the weakest evidence in this article.

The two claims that matter most are insulated from this. The falling share of agent-games matched exactly by one rule, and the rising number of rounds violating the nearest rule, are both computed by exact matching with no learned component at all, and it is precisely those two quantities that the main text uses to argue that play becomes less rule-governed. The compositional claim, which does use the classifier, is reported with its provenance beside it (table~\ref{tab:S-provenance}) and is described in the main text as the weaker of the two.

There is a further and more general caveat. A strategy is a decision rule and not a decision, so it is never observed; separating the memory-one rules from observed play has been shown to require a long run of observations even where the resident rule is fixed by construction~\cite{kim2021win,hilbe2013memory}. Our trajectories are ten rounds long. This is why the read-out is reported throughout as a resemblance and never as an identification, and why richer strategy vocabularies, including zero-determinant and extortionate play~\cite{pressDyson2012}, and mixture estimators designed for longer horizons~\cite{fontana2025nicer,inferenceStrategies}, are named as the natural extension rather than attempted here.

\subsection{Parsing, fallback and the direction of the bias}\label{appendix:robust_parsing}

The parsing fallback records an unresolvable reply as Option~A, which is defection under this framing (\S\ref{appendix:endpoints}). Any contamination it introduces therefore lowers a measured cooperation rate. Two consequences follow. The cooperative behaviour reported in the main text cannot be an artefact of the fallback, because the fallback cannot produce cooperation. And if the fallback rate were correlated with the payoff scale, it would attenuate a measured range rather than create one, so the smallest ranges in table~1 are the ones such a correlation could most plausibly distort, and they are also the ones the article leans on least.

\rev{The per-decision records retained for Gemini 3.5 Flash-Lite allow the rates to be counted directly. In all 50 language-by-scale cells, of 800 decisions each, no reply needed a retry and none fell back to Option~A, so for this model parsing contributes nothing to the scale effect; the per-cell counts are released as \texttt{T27\_parse\_fallback.csv}. For the other five models only the per-game tables were retained, which record the final decision but not the number of attempts, so their retries and fallbacks cannot be reconstructed by language and scale, and the run-level assertion of a fallback rate of at most 0.02 (\S\ref{appendix:endpoints}) is the only bound available.}

\subsection{What this design does not measure}\label{appendix:robust_gaps}

Two limits are worth naming so that they are not mistaken for findings. First, an independent repeat of the Gemini 3.5 arm at $\lam \in \{0.1,0.25,0.5,1\}$ provides a limited test-retest check. The absolute between-run differences in cooperation were 0.002, 0.014, 0.002 and 0.038, respectively. This result supports repeatability for those four conditions but does not provide a reproducibility floor for other models, languages or scales.

Second, clustering on the dyad handles the dependence between the two agents of a game and between the rounds inside an agent-game, but it is not a mixed-effects model with random intercepts for replicate and game, which would be the fuller treatment of a nested design. Because the direction of every effect is consistent across conditions and the clustered intervals are wider than their unclustered counterparts, we regard the reported inference as conservative rather than optimistic, but we do not claim it is complete. The matched-block randomisation provides a complementary sensitivity analysis, not a replacement for a collection record that preserves provider seed metadata.

The release also contains a targeted notation control for Gemini 3.5 Flash-Lite. At fixed payoff magnitude, changing two decimal places to three changed the cooperation rate by $0.0073$ at $\lambda=0.1$ ($p=0.84$) and by $0.011$ at $\lambda=1$ ($p=0.68$). The corresponding magnitude contrast from $0.1$ to $1$ under three-decimal formatting was $0.096$ ($p=0.0018$). This small control does not prove that notation is irrelevant, but it weakens the claim that a single trailing-zero convention explains the main scale effect.

\subsection{Matched scale ladders and influence checks}\label{appendix:robust_ladders}

\rev{A ladder is the set of ten dyads, one per payoff scale, that share a model, a language and a game identifier, and therefore a persona pairing and the requested sampling seeds; each model has 200. The main analyses resample and cluster on the dyad, which treats the ten dyads of a ladder as independent. Table~\ref{tab:D-ladder} repeats the headline inferences with the ladder as the unit. All 18 scale tests remain significant ($p < 2\times10^{-6}$), and the range intervals change little. The persona sign reversal holds in the same two models, Claude and Gemini 3.5, and persistence exceeds reciprocity in every model, narrowly for Claude (0.046, 95\% CI 0.001 to 0.092). Clustering the pooled five-model test on the ladder, and removing each model in turn, leaves the scale effect significant in every fit (largest $p = 4\times10^{-8}$, without Grok).}

\begin{table}[htbp]
\ifhighlightrevisions\color{red}\fi
\centering
\footnotesize
\setlength{\tabcolsep}{2pt}
\caption{\rev{\textbf{Headline inferences with the matched ladder as the unit.} A ladder is the ten dyads, one per payoff scale, that share a model, a language and a game identifier; each model has 200. Intervals are 95\% percentile intervals from 1{,}500 whole-ladder bootstrap draws. The three $\chi^2_9$ statistics test the nine payoff-scale contrasts: a linear model of dyad cooperation with one fixed effect per ladder, a Friedman test on the 200 ladders by ten scales, and the round-level logistic model of table~1 with standard errors clustered on the ladder. Every one of the 18 tests has $p < 2\times10^{-6}$. Persona shift is the persona effect above the sub-unit boundary minus that below it; the last column is the persistence contrast minus the reciprocity contrast.}}
\label{tab:D-ladder}
\begin{tabular}{@{}lcccccc@{}}
\toprule
Model & range & FE $\chi^2_9$ & Friedman $\chi^2_9$ & logit $\chi^2_9$ & persona shift & persist.\,$-$\,recip. \\
\midrule
Claude Haiku 4.5 & $0.081\,[0.062,\,0.127]$ & 43.4 & 45.3 & 48.8 & $0.429\,[0.371,\,0.487]$ & $0.046\,[0.001,\,0.092]$ \\
GPT-5.4 Nano & $0.243\,[0.200,\,0.290]$ & 177.4 & 164.6 & 185.5 & $0.153\,[0.099,\,0.208]$ & $0.135\,[0.107,\,0.162]$ \\
Gemini 3.5 Flash-Lite & $0.190\,[0.169,\,0.242]$ & 168.7 & 214.3 & 141.9 & $0.680\,[0.613,\,0.743]$ & $0.404\,[0.374,\,0.434]$ \\
Qwen3 235B-A22B & $0.095\,[0.069,\,0.123]$ & 58.1 & 78.3 & 85.9 & $0.233\,[0.204,\,0.260]$ & $0.844\,[0.820,\,0.864]$ \\
Grok 4.20 & $0.636\,[0.593,\,0.678]$ & 889.4 & 1059.9 & 977.0 & $0.138\,[0.086,\,0.188]$ & $0.502\,[0.452,\,0.545]$ \\
Gemini 3.1 Flash-Lite & $0.072\,[0.061,\,0.090]$ & 116.8 & 181.7 & 161.0 & $0.103\,[0.088,\,0.118]$ & $0.791\,[0.767,\,0.812]$ \\
\bottomrule
\end{tabular}
\end{table}

The scale labels can also be permuted within each complete game-ID ladder. This is a stricter null than the main analysis because it preserves the recorded matching across all ten scales. With 10,000 blocked permutations, the observed scale range remains significant for every model: $p=0.0019$ for Claude and $p=0.0001$ for each of GPT, Gemini 3.1, Gemini 3.5, Grok and Qwen. The ranges are unchanged because the test changes only the null assignment. The complete results are included in the released analysis tables.

As a complementary influence check, removing each model in turn leaves a significant pooled scale effect. The scale-test $p$-values are below $0.01$ in all six leave-one-model-out fits, with pooled ranges from $0.048$ to $0.159$. Thus the dyad-level pooled association is not removed by any single model, even though the magnitude of the pooled range changes. The complete results are included in the released analysis tables.

\subsection{Uncertainty in the conditioning contrasts}\label{appendix:conditioning_ci}

The main text compares the relative size of partner conditioning and own-action conditioning. To show the uncertainty directly, table~\ref{tab:conditioning-ci} reports percentile intervals from 4,000 resamples of whole dyads. The resampling unit keeps both agents and all ten rounds of a game together. Reciprocity is the partner main effect, persistence is the own-action main effect, and matching is their interaction. The intervals confirm that persistence is positive in every model, whereas reciprocity is positive in four models and negative in Gemini 3.1 and Qwen. These are descriptive contrasts and do not identify a causal mechanism.

The persistence comparison is stable to the context-count rule. Across minimum context thresholds of 10, 20, 30, 40 and 50 observations, the mean persistence-minus-reciprocity contrast remains positive for every model. A Jeffreys-smoothed estimate using all observed contexts gives the same result. The complete threshold grid is included in the released analysis tables. The result is strongest for Gemini 3.1, Gemini 3.5, Grok and Qwen; Claude's average difference is small and some individual cells are close, so we interpret the comparison in the heading of main-text section~3.7 as a relative corpus-level tendency rather than a universal cell-level rule.

Finally, a round-level logistic model with scale, language, both persona terms and their scale-by-language interactions rejects the joint null of no scale-by-language interaction for every model. The largest Holm-adjusted $p$-value is $1.4\times10^{-48}$, so the interaction is not only a visual range difference. Its interpretation remains conditional on the language templates used here, including the documented Arabic and Chinese wording difference. Full coefficients and adjusted tests are included in the released analysis tables.

\begin{table}[htbp]
\centering
\footnotesize
\setlength{\tabcolsep}{4pt}
\caption{\textbf{Whole-dyad bootstrap intervals for memory-one contrasts.} Entries are estimates with 95\% percentile intervals. Each model contributes 2,000 dyads and each interval uses 4,000 bootstrap draws.}
\label{tab:conditioning-ci}
\begin{tabular}{@{}lccc@{}}
\toprule
Model & reciprocity & persistence & matching \\
\midrule
Claude Haiku 4.5 & $0.243\,[0.227,\,0.258]$ & $0.288\,[0.273,\,0.303]$ & $0.172\,[0.159,\,0.186]$ \\
GPT-5.4 Nano & $0.141\,[0.127,\,0.156]$ & $0.277\,[0.263,\,0.291]$ & $0.053\,[0.040,\,0.066]$ \\
Gemini 3.5 Flash-Lite & $0.169\,[0.153,\,0.185]$ & $0.573\,[0.555,\,0.590]$ & $0.020\,[0.006,\,0.035]$ \\
Qwen3 235B-A22B & $-0.095\,[-0.107,\,-0.083]$ & $0.749\,[0.735,\,0.762]$ & $-0.085\,[-0.096,\,-0.074]$ \\
Grok 4.20 & $0.189\,[0.175,\,0.203]$ & $0.690\,[0.676,\,0.704]$ & $0.007\,[-0.006,\,0.019]$ \\
Gemini 3.1 Flash-Lite & $-0.171\,[-0.185,\,-0.158]$ & $0.620\,[0.606,\,0.633]$ & $0.168\,[0.157,\,0.179]$ \\
\bottomrule
\end{tabular}
\end{table}

\section{Reproduction}\label{appendix:repro}

Every number, figure and table in this article and in this appendix is produced by an archived pipeline from the released transcript corpus. Tables and figures are generated directly; numerical values quoted manually in prose are checked against the data by the verification ledger. The three main-text tables are generated into \texttt{tables\_auto.tex}, the supplementary tables of \S\ref{appendix:results} into \texttt{supp\_tables\_auto.tex}, and the evolutionary grid of \S\ref{appendix:egt} into \texttt{egt\_tables\_auto.tex}.

The pipeline runs in the following order. \texttt{s00\_build} reads the transcript corpus, applies the completeness guard of \S\ref{appendix:corpus} and writes the tidy game, round and read-out tables. \texttt{s01\_train\_lstm} trains the classifier of \S\ref{appendix:classifier} on the synthetic corpus. \texttt{s02\_readout} applies the rule base and the classifier to the transcripts. \texttt{s03\_stats} and \texttt{s04\_strategy\_stats} compute every statistic reported here. The complete artifact is rebuilt by \texttt{Analysis/\allowbreak scaling/\allowbreak rebuild\_interface\_focus.py}. This driver encodes the dependency order: deterministic data assembly and read-out, primary statistics, matched-ladder and reviewer robustness ledgers, evolutionary calculations, the current figure modules in \texttt{Analysis/figures}, generated tables and final verification. It deliberately does not call the legacy \texttt{s05} figure drivers. The verifier recomputes headline quantities quoted in the manuscript prose from the data and fails if any of them disagrees, so that a stale number cannot survive an edit.

The collection uses base sampling seed 12345 where supported by the serving interface. Classifier training uses seed 1234. Primary bootstrap and permutation analyses use 20260905, descriptive figure resampling uses 20260906, and the matched-block and conditioning robustness analyses use 20260907. Provider-specific seed semantics are not assumed.

The results reported here were produced under Python 3.12.3 with NumPy 1.26.4, pandas 2.2.2, PyArrow 16.1.0, SciPy 1.17.1, statsmodels 0.14.6, scikit-learn 1.5.1, PyTorch 2.11.0 on CPU, Matplotlib 3.9.2, mpmath 1.3.0 and EGTtools 0.1.14.2. The classifier is trained on CPU and its seed is fixed, so the read-out is reproducible without a GPU.

\rev{\section{Supplementary discussion}\label{appendix:discussion}

\subsection{Comparison with human stake experiments}\label{appendix:human}

In human experiments, the effect of raising real stakes is usually modest and depends on the task: contributions in a public goods game change little when stakes are raised~\cite{kocher2008does}, dictators and ultimatum proposers give a slightly smaller share as stakes grow~\cite{engel2011dictator,larney2019stake}, responders reject fewer offers at very high stakes~\cite{cameron1999raising,andersen2011stakes}, and risk aversion rises with the scale of real payoffs~\cite{holt2002risk}. Higher incentives more often reduce the variance of behaviour than shift its mean~\cite{camerer1999effects}.

The departures reported in the main text differ from these effects in three ways. They differ in size: in human experiments even large changes in real money move behaviour modestly, whereas here a change of units with no consequence for anyone moves cooperation by up to 0.636 (Grok). They differ in direction: for a given game, human stake effects tend to point the same way across studies, whereas here the five models move in different and often opposite directions. They differ in location: human stake effects appear when stakes become large relative to income, whereas here much of the change occurs at or below $\lam = 1$, a range in which no human participant would perceive a meaningful difference in stakes. This argues against the simplest account, in which the models reproduce a human response to stakes learned from text, although it does not rule out a distorted version of it.

\subsection{Why the models might differ}\label{appendix:hypotheses}

We see four explanations, which are not mutually exclusive; each is consistent with part of the evidence and each can be tested with a targeted design.

First, the prompt contains components that do not scale with $\lam$: the narrative framing, the persona line and the goal sentence. Rescaling changes the weight of the numbers relative to this fixed text, and models may differ in how strongly they attend to each part. This is consistent with the persona effect depending on the payoff scale and reversing its sign at sub-unit stakes in Claude and Gemini (main text, Section~3.4). Removing the narrative or the persona while keeping the matrix would test it.

Second, the numbers themselves are represented differently: sub-unit payoffs are rare in reward contexts and are split into different tokens from integers. This is consistent with much of the change occurring at or below $\lam = 1$. The notation control of \S\ref{appendix:robust_gaps}, however, moved cooperation by about 0.01 at fixed magnitude, against 0.096 for the magnitude contrast, so a single formatting convention does not explain the effect. Varying the format more widely at fixed magnitude, for example in words or in scientific notation, would test this account.

Third, providers differ in post-training, including instruction tuning and preference optimisation, which shapes how a model follows instructions. Qwen and Grok invert the persona instruction at every scale (main text, Section~3.4), which points to such differences. Comparing the base and instruction-tuned versions of one open-weight model would isolate this source.

Fourth, the models may reproduce associations between magnitude and importance learned from human text. The comparison in \S\ref{appendix:human} suggests that, if they do, the learned response is distorted. The verbal baseline of \S\ref{appendix:verbal} would show how much of the effect requires numbers at all.

\subsection{A verbal baseline}\label{appendix:verbal}

The design has no condition without numbers. A verbal baseline, in which each outcome is described only by its rank (for example, that a player who defects alone receives the best possible outcome), would keep the ordinal structure of the game while removing every magnitude. It would show whether the presence of numerical payoffs itself changes behaviour, and it would give an empirical reference against which each payoff scale could be compared. Running it requires a new collection over all models, languages and personas, and is a natural next step.}

\end{document}